\documentclass{article}

\usepackage[preprint]{neurips_2026}

\usepackage[utf8]{inputenc} \usepackage[T1]{fontenc}    \usepackage{hyperref}       \usepackage{url}            \usepackage{booktabs}       \usepackage{amsfonts}       \usepackage{nicefrac}       \usepackage{microtype}      \usepackage{xcolor}         \usepackage[dvipsnames]{xcolor}
\usepackage{enumitem}

\usepackage{graphicx}
\usepackage{subcaption}
\usepackage{float}
\usepackage{tabularx}
\usepackage{multirow}
\usepackage{adjustbox}
\usepackage{tikz}

\usepackage{amsmath}

\newcommand{\R}{\mathbb{R}}

\usepackage[normalem]{ulem} 

\title{LLM Pretraining Shapes a Generalizable Manifold: Insights into Cross-Modal Transfer to Time Series}

\author{
  Alexis Roger* \\
  McGill University \\
  Mila - Quebec AI Institute \\
  \And
  Prateek Humane* \\
  Université de Montréal \\
  Mila - Quebec AI Institute \\
  \And
  Zhenghan Tai \\
  University of Toronto \\
  \And
  Gwen Legate \\
  Concordia University \\
  Mila - Quebec AI Institute \\
  \And
  Andrei Mircea \\
  Université de Montréal \\
  Mila - Quebec AI Institute \\
  \And
  Vasilii Feofanov \\
  42.com \\
  \And
  Irina Rish \\
  Université de Montréal \\
  Mila - Quebec AI Institute
}

\begin{document}
\maketitle

\begin{abstract}
Can language-pretrained transformers become effective time-series forecasters, and why? 
In this paper, we show that cross-modal transfer arises because language pretraining preconditions time series training with a reusable manifold. 
A linear probe on frozen LLM states decodes realistic time-series trajectories without paired supervision, and retrieval in this projected space yields competitive forecasts, showing that structure and dynamics exist before finetuning. 
Pretrained initialization also improves optimization, producing coherent gradients and a highly anisotropic loss landscape unlike random initialization. 
Finetuning then acts as low-dimensional alignment, reusing existing directions rather than learning temporal primitives from scratch, as evidenced by low-rank updates, subspace alignment, and shared features for periodicity, trend, and repetition. 
Together, these results support a geometric account of LLM-to-time-series transfer: language pretraining builds the manifold, and finetuning projects numerical dynamics onto task-relevant directions.
\end{abstract}

\begin{figure}[h]
    \centering
    \includegraphics[width=0.85\linewidth]{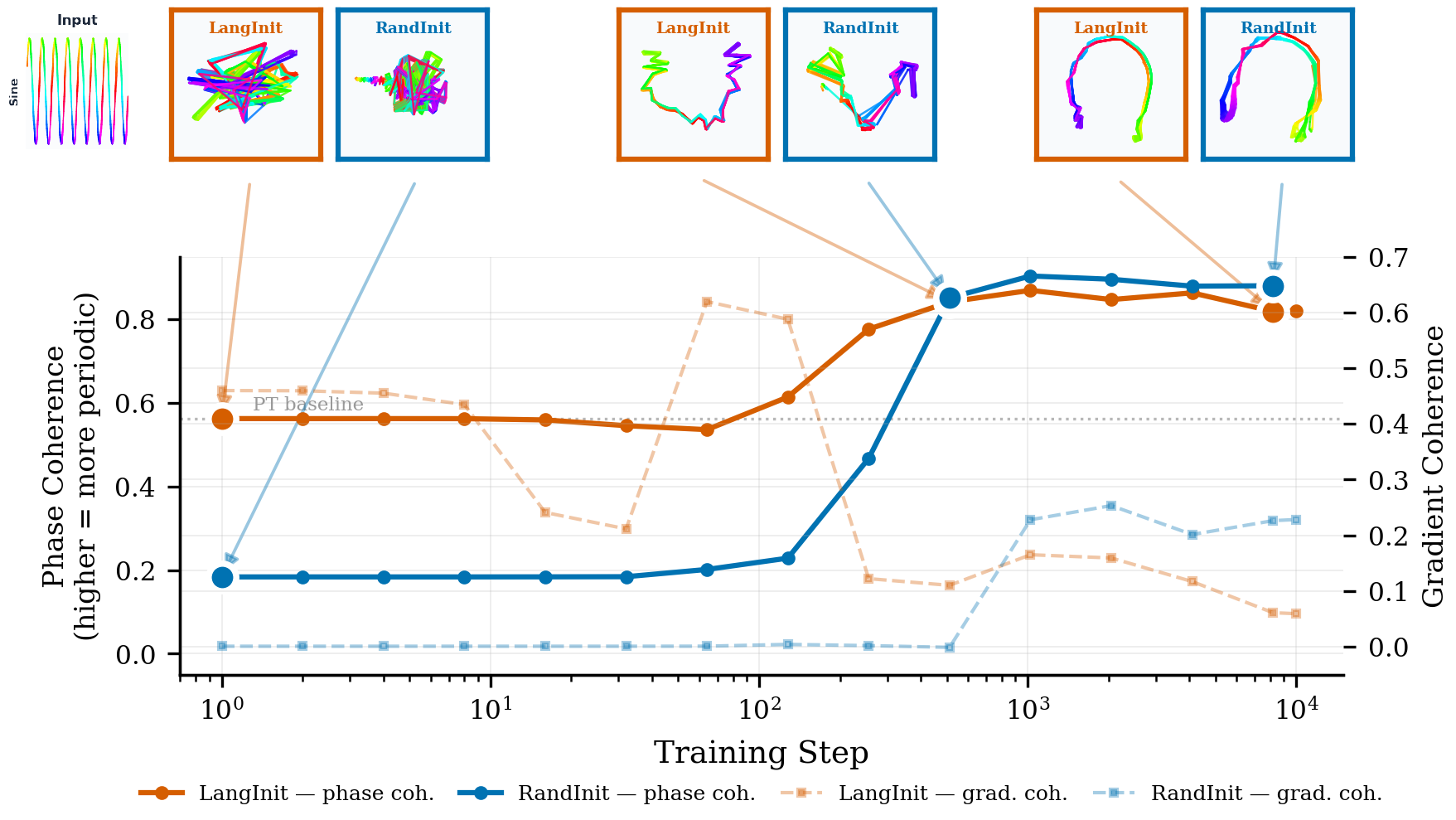}
    \caption{\textbf{Phase and gradient coherence across training.} 
Higher the coherence, better the model captures the periodic nature of the input (a sine wave here). Using t-SNE, we visualize hidden states at training steps 1, 512, and 8192 and see that pretrained LLM (LangInit) inherits periodic structure from pretraining, starting high. Random initialization (RandInit) starts near zero and undergoes an abrupt phase transition around step 300 where gradient coherence spikes, indicating that the model discovers periodic geometry. This is followed by aligning its gradients across the time series.}
    \label{fig:coherence_evolution}
\end{figure}

\section{Introduction}
\vspace{-0.3cm}

Large Language Models (LLMs) transfer effectively across many downstream tasks, but it remains unclear whether their pretrained representations generalize beyond language itself. Time series forecasting provides a compelling test case: like language modeling, it is a sequential prediction problem, yet it operates on continuous numerical signals rather than discrete semantic tokens. If language pretraining improves forecasting, this would suggest that transformers learn reusable sequential structure rather than purely linguistic knowledge.

Recent work has explored routes from LLMs to time series, including input reprogramming~\citep{jin2024timellmtimeseriesforecasting}, zero-shot digitized prompting~\citep{gruver2024largelanguagemodelszeroshot, williams2025contextkeybenchmarkforecasting}, parameter-efficient finetuning~\citep{zhou2023fitsallpowergeneraltime, wolff2025usingpretrainedllmsmultivariate}, and tokenized forecasting approaches such as Chronos~\citep{ansari2024chronoslearninglanguagetime, talukder2025totemtokenizedtimeseries, roger2025smallvocabulariesbiggains}. However, empirical evidence remains contradictory. Some studies report little benefit from language pretraining and attribute gains primarily to transformer architectures or tokenization effects~\citep{ansari2024chronoslearninglanguagetime, tan2024are, zheng2025, zhang2025from}, while others find strong improvements in low-data, cross-domain, or distribution-shift settings~\citep{riachi2025randominitializationcantcatchup, bayazi2024general, qiu2026rethinkingrolellmstime}. 

We believe that the transferred object is not semantic knowledge, but representational geometry. Language contains rich temporal structure (e.g., repetition, trends, periodicity, and long-range dependencies) and autoregressive pretraining organizes transformers into a sequential manifold encoding such dynamics. Under this view, time-series finetuning becomes a low-dimensional alignment problem: adapting existing temporal directions rather than learning forecasting structure from scratch.

To test this hypothesis, we repurpose Qwen3-0.6B \citep{qwen3technicalreport}, using next-token prediction paradigm for probabilistic time-series forecasting. We compare pretrained and randomly initialized models across full finetuning and parameter-efficient adaptation strategies on GiftEval benchmark \citep{aksu2024giftevalbenchmarkgeneraltime}. Our results show that pretrained models converge faster than identical architectures trained from scratch (Figure \ref{fig:coherence_evolution}). We further find that pretrained hidden states linearly decode realistic temporal trajectories, produce coherent optimization gradients, and adapt through low-rank updates, supporting a geometric account of cross-modal transfer. Together, these findings suggest that LLMs help forecasting not because they understand the semantics of a signal, but rather because language pretraining already organizes the model into a reusable temporal representation space.
\vspace{-0.1in}
\section{Related Work}
\vspace{-0.1in}
\textbf{LLMs and time-series forecasting.}
Recent time-series foundation models either pretrain directly on temporal data~\citep{rasul2024lagllamafoundationmodelsprobabilistic, goswami2024momentfamilyopentimeseries,woo2024unifiedtraininguniversaltime} or adapt pretrained LLMs through reprogramming, frozen-backbone alignment, or parameter-efficient finetuning~\citep{jin2024timellmtimeseriesforecasting, zhou2023fitsallpowergeneraltime, gruver2024largelanguagemodelszeroshot, wolff2025usingpretrainedllmsmultivariate}. Tokenization-based approaches discretize continuous signals into vocabularies and apply language-model objectives~\citep{ansari2024chronoslearninglanguagetime, talukder2025totemtokenizedtimeseries, roger2025smallvocabulariesbiggains}. While some studies argue that forecasting gains stem mainly from architectural biases or tokenization effects~\citep{tan2024are, zheng2025, zhang2025from}, more recent large-scale evaluations show that language pretraining becomes particularly beneficial under distribution shift and cross-domain generalization~\citep{qiu2026rethinkingrolellmstime}. Our work complements this literature by studying not only \emph{when} transfer occurs, but \emph{why} it occurs.

\textbf{Cross-modal representation geometry.} The Platonic Representation Hypothesis~\citep{huh2024platonicrepresentationhypothesis} posits that sufficiently scaled models converge toward a shared statistical structure of reality, regardless of modality. Supporting this, embeddings exhibit universal cross-modal geometry~\citep{jha2025harnessinguniversalgeometryembeddings}, and frozen vision backbones can forecast time series~\citep{chen2024visionts,roschmann2025time}. These findings motivate our hypothesis that inherent temporal structures already exist within language-pretrained transformers.

\textbf{Loss geometry and low-dimensional adaptation.}
Prior work shows that optimization geometry plays a central role in transfer learning with analysis of loss landscapes~\citep{li2018visualizinglosslandscapeneuralnets, ghorbani2019investigationneuralnetoptimization, fort2019largescalestructureneural} and efficient finetuning of pretrained models within low-dimensional subspaces~\citep{aghajanyan-etal-2021-intrinsic} such as LoRA~\citep{hu2021loralowrankadaptationlarge}. We extend these ideas to cross-modal transfer, showing that language-pretrained transformers adapt to time series via coherent gradients and low-rank specialization rather than full representational rewriting.

\vspace{-0.1in}
\section{Training Methodology}
\label{sec:transfer}
\label{sec:training_method}
\vspace{-0.1in}

We repurpose Qwen3-0.6B~\citep{qwen3technicalreport} as a probabilistic time-series forecaster by casting forecasting as next-token prediction over discretized values. 
\vspace{-0.1in}
\subsection{Tokenization and Forecasting}
\vspace{-0.1in}
We extract sliding windows of length $C+L$, with context length $C=512$ and prediction horizon $L\!=\!64$. Each series is normalized using statistics from the context window and discretized into $V\!=\!1024$ uniform bins over $[-5,5]$. Following~\citet{roger2025smallvocabulariesbiggains}, bin indices are mapped directly to the first vocabulary positions of the pretrained model, while standard Qwen3 special tokens are preserved.
The model is trained using a weighted quantile loss over quantile levels $\mathcal{Q}=\{0.1,\dots,0.9\}$, following Chronos Bolt~\citep{ansari2024chronoslearninglanguagetime}:
\begin{equation}
\label{eq:quantile_loss}
\mathcal{L} = \frac{1}{T \cdot |\mathcal{Q}|} \sum_{t=1}^{T} \sum_{\tau \in \mathcal{Q}} \rho_\tau\!\left(y_t - \hat{q}_{t,\tau}\right), \qquad \rho_\tau(u) = u \cdot \left(\tau - \mathbf{1}[u < 0]\right),
\end{equation}
where $\rho_\tau$ is the standard quantile regression loss. Output logits are transformed into a categorical distribution over ordered bins, defining a discrete CDF from which arbitrary quantiles are extracted by inversion. This produces coherent non-crossing probabilistic forecasts from a single forward pass, without requiring quantile-specific training or repeated inference.
\vspace{-0.1in}
\subsection{Experimental Setup}
\label{sec:training_regimes}
\vspace{-0.1in}
Models are trained on sliding windows from the GiftEval corpus~\citep{aksu2024giftevalbenchmarkgeneraltime} using AdamW with linear warmup and decay. Training uses an effective batch size of 128 on NVIDIA A100 GPUs with bf16 precision. Evaluation is performed on 1,000 held-out windows using CRPS, MASE, and MSE amongst others error metrics following GluonTS conventions (see Appendix \ref{sec:appendix_metrics}).

For a pretrained LLM (Qwen3-0.6B), we compare four adaptation strategies that differ only in trainable parameters:
\vspace{-0.4em}
\begin{itemize}[leftmargin=1em, itemsep=0pt, topsep=2pt, parsep=0pt]
    \item \textbf{Full finetuning:} all model parameters are updated.
\item \textbf{IO Only:} transformer backbone is frozen, only the input embeddings and output head are trained.
\item \textbf{LoRA Attn:} LoRA adapters 
are applied only to attention projections, the rest is frozen.
\item \textbf{LoRA Attn+IO:} LoRA attention adaptation with trainable input embeddings and output head.
\end{itemize}

\vspace{-0.1in}
\subsection{Results: Transfer from Language to Time Series}
\begin{figure}
    \centering
    \includegraphics[width=\linewidth]{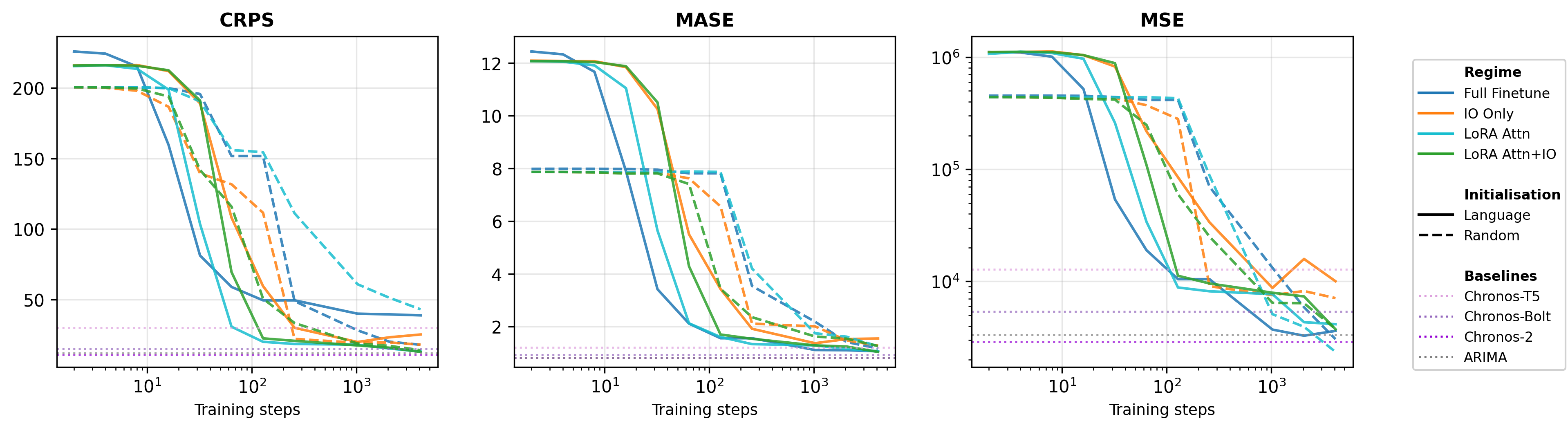}
    \caption{Training progression of h=1 forecasting metrics across training steps for four training regimes. Solid lines denote language-pretrained Qwen3-0.6B initialization; dashed lines denote random initialization. Horizontal dotted lines indicate baseline model performance. Language-pretrained models converge faster and to lower error across all regimes, with LoRA-based methods matching full finetuning performance.}
    \label{fig:small_h1_progression}
\end{figure}

\vspace{-0.1in}
Figure~\ref{fig:small_h1_progression} (as well as  Figure 6 in section \ref{sec:finetuning}) shows that language-pretrained models consistently converge faster and reach lower forecasting error than randomly initialized counterparts across all adaptation regimes. The advantage is strongest during early optimization, particularly for CRPS and MASE, indicating that language pretraining provides a favorable sequential inductive bias. LoRA-based adaptation achieves performance comparable to full finetuning, supporting the hypothesis that forecasting transfer operates through low-dimensional specialization of pretrained representations rather than complete relearning.  These observations hold on longer horizons, including h=64 which can be found in Appendix \ref{sec:appendix_experimental_setup}, along with the complete set of metrics.

\vspace{-0.1in}
\section{The Shared Manifold Hypothesis}
\label{sec:prior}
\vspace{-0.1in}

We now present a geometric explanation for the observed phenomena. Due to the recursive structure and statistical patterns exhibited by language \citep{ou2025identifying}, we theorize that LLM representations already contain native temporal structures well suited for adaptation to time series data. We present several experiments demonstrating that LLMs do possess a manifold that is already compatible with time series forecasting, and that finetuning does not need to create new directions, it simply identifies and collapses time series representations onto the best pre-existing directions.
\vspace{-0.1in}
\subsection{A Linear Probe Reveals Latent Temporal Structure}
\label{sec:linear_decoding}
\vspace{-0.1in}
\begin{figure}[!t]
\centering
\begin{subfigure}[t]{0.45\textwidth}
\centering
\includegraphics[width=\textwidth]{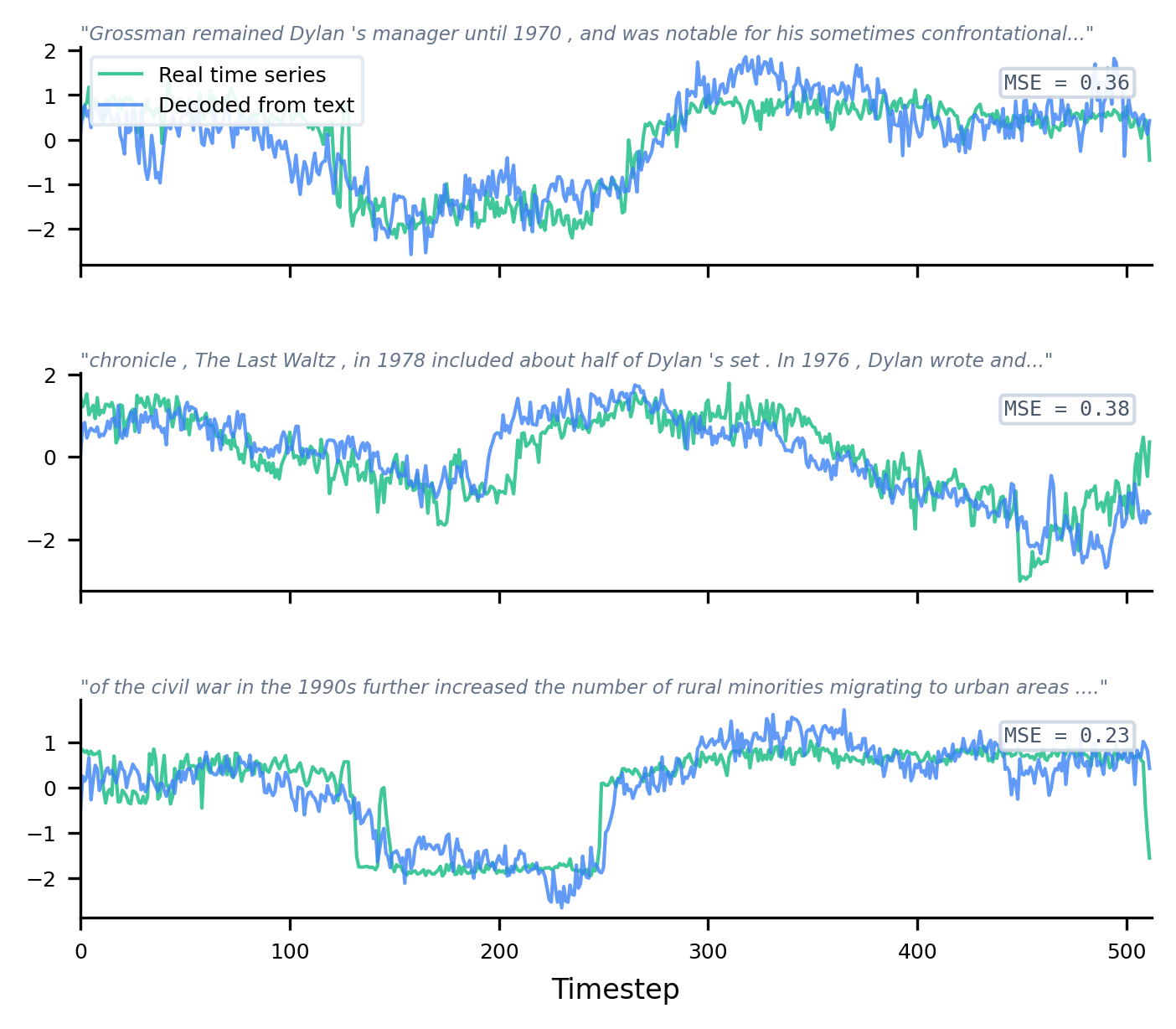}
\caption{Decoded time series from text}
\label{fig:examples}
\end{subfigure}
\hfill
\begin{subfigure}[t]{0.52\textwidth}
\centering
\includegraphics[width=\textwidth]{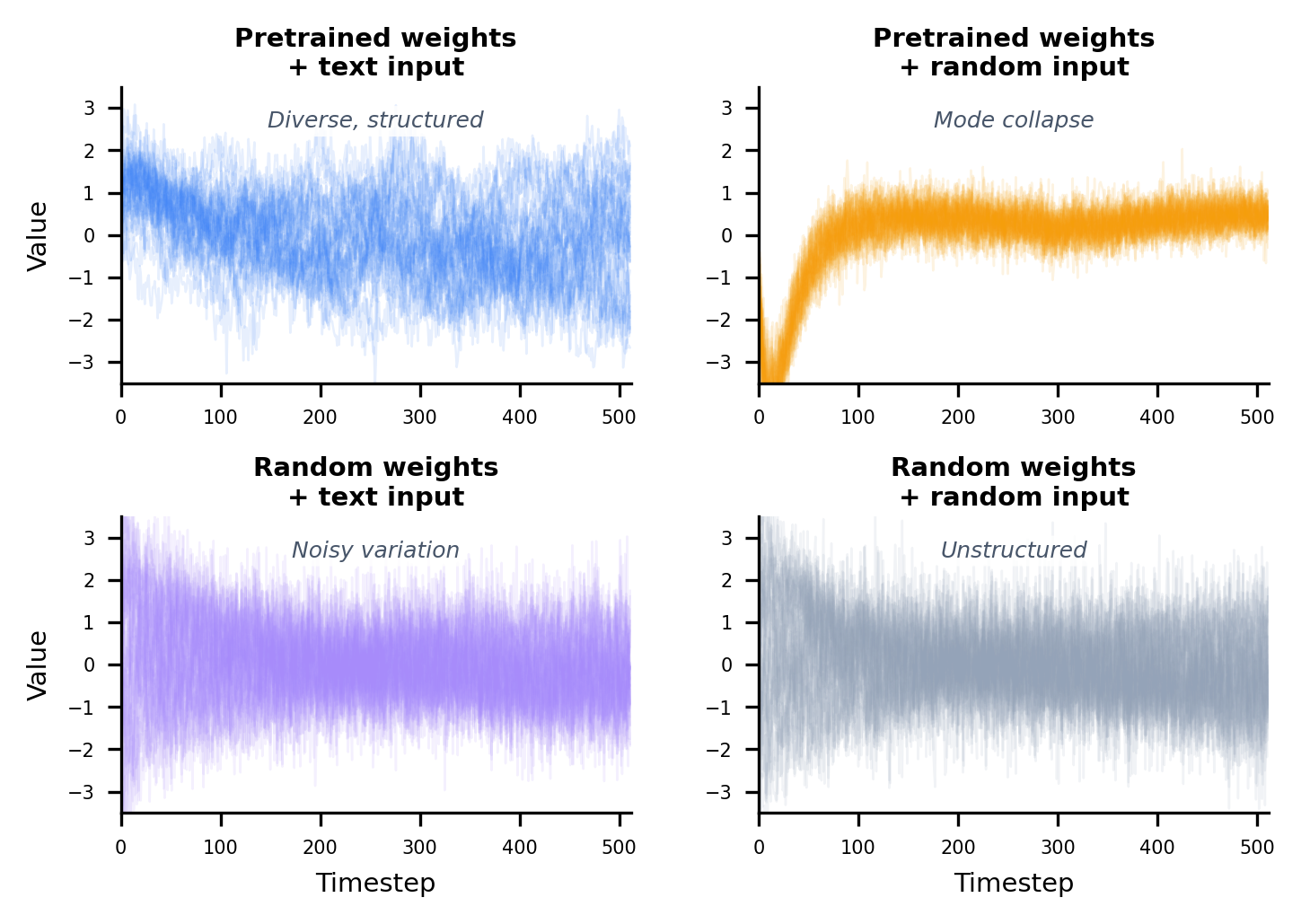}
\caption{Ablation: what creates the structure?}
\label{fig:ablation}
\end{subfigure}
\caption{\textbf{Pretrained hidden states contain time-series-compatible structure.} A single linear map $\hat{y}_t = \mathbf{w}^\top \mathbf{h}_t + b$ is trained on frozen LLM hidden states via EM-style nearest-neighbor matching to a bank of 10{,}000 real time series (no paired data). \textbf{(a)} Three decoded outputs (blue) overlaid with their nearest real time series (green). Input text shown above each plot (gray). MSE $= 0.23$--$0.38$; full distribution in Appendix. \textbf{(b)} Ablation crossing learned weights vs.\ random init with text vs.\ random input. \emph{Top-left}: pretrained + text---diverse, structured outputs. \emph{Top-right}: pretrained + random tokens---mode collapse to one shape (manifold exists but is not traversed). \emph{Bottom-left}: random weights + text---noisy variation without structure. \emph{Bottom-right}: random weights + random tokens---unstructured. Only pretrained weights \emph{and} meaningful text together produce diverse temporal signals.}
\label{fig:hero}
\end{figure}

We test if the distribution of pretrained language model representations is already geometrically compatible with the distribution of real time series: can a simple linear map, trained without paired text--time-series supervision, project frozen LLM hidden states onto realistic time-series windows?

\textbf{Method. } We pass $N = 1{,}920$ WikiText-103 sequences of $T = 512$ tokens through the frozen Qwen3-0.6B model. For each token, we represent the LLM state by concatenating the hidden states from all 28 layers:
\begin{equation}
    \mathbf{h}_{i,t} = \bigl[\mathbf{h}_{i,t}^{(0)};\; \mathbf{h}_{i,t}^{(1)};\; \ldots\;;\; \mathbf{h}_{i,t}^{(27)}\bigr] \in \R^{D}, \quad D = 28 \times 1024 = 28{,}672.
\end{equation}

A single linear layer is then applied independently at each timestep to produce a scalar prediction:
\begin{equation}
\label{eq:linear_map}
    \hat{y}_{i,t} = \mathbf{w}^\top \mathbf{h}_{i,t} + b, \quad \mathbf{w} \in \R^D,\; b \in \R.
\end{equation}
This produces a predicted sequence $\hat{\mathbf{y}}_i \in \R^T$ for each input text. The output is z-score normalized to zero mean and unit variance, matching the normalization of the target time series.

\textbf{Training objective. } We use an EM-style algorithm to match text and time-series targets. We match each prediction to its nearest z-scored window from a bank of $M = 10{,}000$ GiftEval time series \citep{aksu2024giftevalbenchmarkgeneraltime}. At each step, we sample $K = 128$ candidate windows per prediction, choose $j_i^* = \arg\min_{j} \frac{1}{T}\|\hat{\mathbf{y}}_i - \mathbf{Y}_j\|^2$, and optimize:
\begin{equation}
\label{eq:loss}
    \begin{aligned}
    \mathcal{L} &=
    \underbrace{\frac{1}{B}\sum_{i=1}^{B} \frac{1}{T}\|\hat{\mathbf{y}}_i - \mathbf{Y}_{j_i^*}\|^2}_{\text{MSE to nearest real TS}}
    \quad + \quad \lambda\,
    \underbrace{\frac{1}{|\mathcal{P}|}\sum_{(i,j)\in\mathcal{P}}
    \cos\,\bigl(|\text{FFT}(\hat{\mathbf{y}}_i)|^2,\, |\text{FFT}(\hat{\mathbf{y}}_j)|^2\bigr)}_{\text{PSD diversity penalty}}.
    \end{aligned}
\end{equation}
The nearest-neighbor assignments are treated as fixed targets for the gradient step. The PSD diversity penalty discourages mode collapse; we set $\lambda = 0.5$ and train with Adam ($\eta = 10^{-3}$) for 100 epochs with batch size $B = 32$.

\textbf{Results. } We evaluate each decoded output by finding its nearest neighbor (by per-timestep MSE) among the 10{,}000 real time series in the bank. Across the 1{,}920 training predictions, the nearest neighbors span 686 distinct real time series---36\% of predictions map to a unique real time series rather than collapsing to repeated patterns (Fig.~\ref{fig:examples}). When predictions are ranked by match quality and compared at equal diversity levels, the pretrained model consistently achieves the lowest MSE across all conditions (Appendix~\ref{app:fair_comparison}). On 1{,}000 held-out WikiText sequences never seen during training, the map achieves comparable quality (mean nearest-neighbor MSE 0.72 vs.\ 0.67 on training text), confirming that the temporal structure is a general property of the pretrained representation space.

\textbf{What creates this structure? } We disentangle the contributions of the learned weights, the input, and the architecture with a $2 \times 2$ ablation crossing model weights (pretrained vs.\ randomly initialized, same architecture) with input (WikiText vs.\ random tokens from the full vocabulary). The ablation is decisive: with random initialization, unique coverage drops to 2.8\% (54 distinct matches); with random tokens through pretrained weights, it collapses to 0.2\% (4 distinct matches). Neither the architecture alone nor the input alone is sufficient---only the combination of pretrained weights and meaningful text produces diverse, realistic temporal signals (Fig.~\ref{fig:ablation}).

\vspace{-0.1in}
\subsection{The Projected Dynamics Forecast Beyond the Context}
\label{sec:forecasting}
\vspace{-0.1in}

The linear decoding result shows shape-level similarity between projected hidden states and real time series. We test whether this extends to \emph{predictive} structure: given the first half of a time series, can a retrieved text projection forecast the second half?

\textbf{Method. } We take 500 real time series from GiftEval and observe only the first 256 of 512 timesteps. Among 10{,}000 WikiText projections, we retrieve the one with lowest MSE on the observed first half and use that projection's second half as the forecast. The baseline is last-value carry-forward, \textit{i.e.}\ repeating the final observed value.

\begin{figure}[h]
\centering
\includegraphics[width=\textwidth]{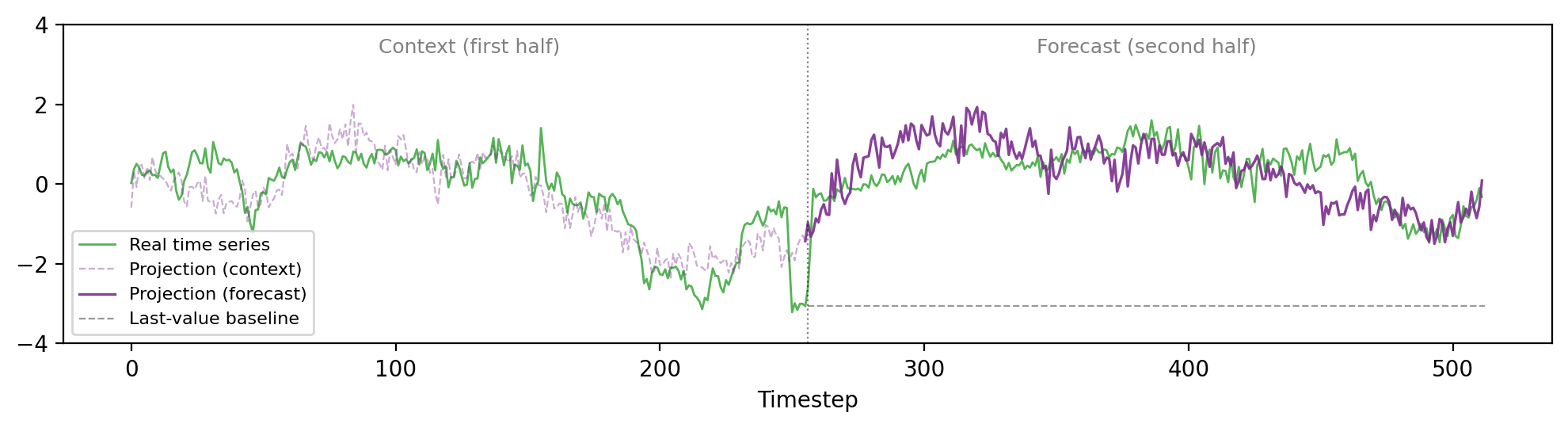}
\caption{Forecasting example with largest gap compared to baseline from 500 evaluation queries. The first 256 timesteps are used to retrieve the closest WikiText projection; the projection's second half (purple) forecasts the continuation. The retrieved projection captures the level shift and subsequent trend (MSE $= 0.42$), while last-value carry-forward (gray dashed) is stranded at the final context value (MSE $= 11.7$).}
\label{fig:forecasting}
\end{figure}

\textbf{Result. } Across all 500 queries, the retrieval-based forecast achieves MSE $= 1.91$ versus last-value's $2.27$ ($16\%$ lower). The retrieval wins in only $37\%$ of cases; however, when it wins, the advantage is large (median $4\times$ lower MSE), because the retrieved projection captures trends and level shifts that last-value misses entirely. Figure~\ref{fig:forecasting} shows the best-case example of this phenomena from the evaluation set.

\vspace{-0.1in}
\subsection{Implications for Transfer}
\label{sec:implications}
\vspace{-0.1in}
The linear probing and forecasting results suggest a possible explanation for why finetuning can succeed with limited adaptation. A pretrained autoregressive language model can be viewed as inducing a deterministic update over its prefix representations:
\begin{equation}
\label{eq:deterministic_update}
    \mathbf{h}_{t+1} = F(\mathbf{h}_t, x_{t+1}),
\end{equation}
where $F$ denotes the transformer's forward computation and $x_{t+1}$ is the next input token. Next-token prediction encourages these hidden states to preserve information relevant to the continuation of the sequence.

Our experiments identify a direction $\mathbf{w}$ whose projected trajectory $\{y_t = \mathbf{w}^\top \mathbf{h}_t\}_{t=1}^T$ resembles real time series (Sec.~\ref{sec:linear_decoding}), and show that these projections contain useful continuation signal in a retrieval setting (Sec.~\ref{sec:forecasting}). Taken together, this suggests that parts of the pretrained representation space are already compatible with time-series-like trajectories, even before any time-series supervision.

Finetuning for time series may therefore be \textit{less about learning temporal behavior from scratch and more about alignment: identifying directions in the existing representation space that are useful for temporal prediction, and adjusting the model to use them more reliably.}

\vspace{-0.1in}
\subsection{Pretraining Creates Coherent Gradient Signal for the TS Objective}
\label{sec:curvature}
\vspace{-0.1in}

\citet{mircea-etal-2025-training} attribute slow training progress in language models to \emph{zero-sum learning}: per-example gradients that conflict, so reducing loss on one subset of examples raises it on another. Here we show that language pretraining resolves this opposition for the time-series objective: pretrained models exhibit coherent, mutually reinforcing gradients across diverse time-series inputs from the outset of finetuning, which bypasses the destructive interference phase that randomly initialized models must first overcome.

\begin{figure}[h]
\label{fig:gradient_alignment}
\centering
\includegraphics[width=\textwidth]{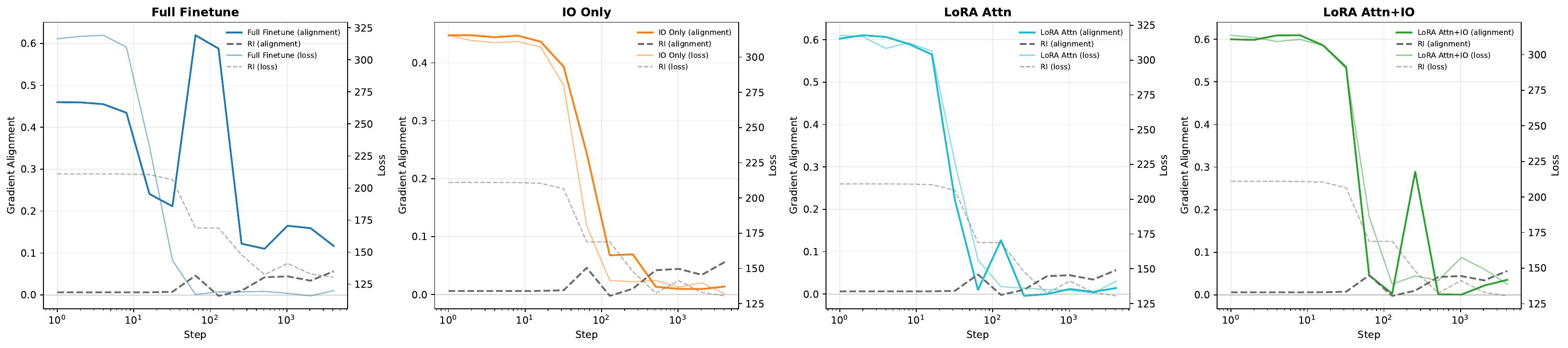}
\caption{\textbf{Gradient alignment and evaluation loss across training for four adaptation regimes.} Each panel shows per-sample gradient alignment (left axis; mean pairwise cosine similarity of individual time-series gradients over 32 held-out examples) and CRPS evaluation loss (right axis) as a function of training steps (log scale). Solid lines denote language-pretrained (LangInit) models; dashed lines denote randomly initialized (RandInit) model. Across all regimes, language-pretrained models exhibit high gradient alignment from step one and loss decreases immediately, whereas randomly initialized models remain near-zero alignment until step~64--256, at which point loss begins to fall, indicating that coherent gradient signal is a prerequisite for effective optimization, and that language pretraining provides it from the outset. } 
\label{fig:grad_alignment}
\end{figure}

\textbf{Method. } For a fixed evaluation batch of $N{=}32$ individual time-series sequences, we compute the per-sample gradient $\mathbf{g}_i = \nabla_\theta \mathcal{L}(x_i;\theta)$ over all model parameters at each checkpoint. We measure gradient alignment as the mean off-diagonal pairwise cosine similarity: 
$
\frac{1}{N(N{-}1)} \sum_{i \neq j} \cos(\mathbf{g}_i,\, \mathbf{g}_j). 
$

\textbf{Interpretation. } 
In Figure \ref{fig:grad_alignment}, we see that across all finetuning regimes, gradient alignment is quite high, especially when compared to gradient alignment in the randomly initialized model. The randomly initialized model takes a warmup phase before it is able to learn a representation such that gradients are aligned and there is a clear learning signal. By this point, the loss has already dropped for the language initialized model. The initial alignment indicates that there is a clear gradient signal to repurpose in pretrained representations and transfer to time series forecasting.

\vspace{-0.1in}
\section{Finetuning Projects the Pretrained Manifold onto Time Series}
\label{sec:finetuning}
\vspace{-0.1in}

\subsection{The Weight Changes are Low-Rank}
\label{sec:lora}
\vspace{-0.1in}

\begin{figure}[h]
    \centering
    \includegraphics[width=\linewidth]{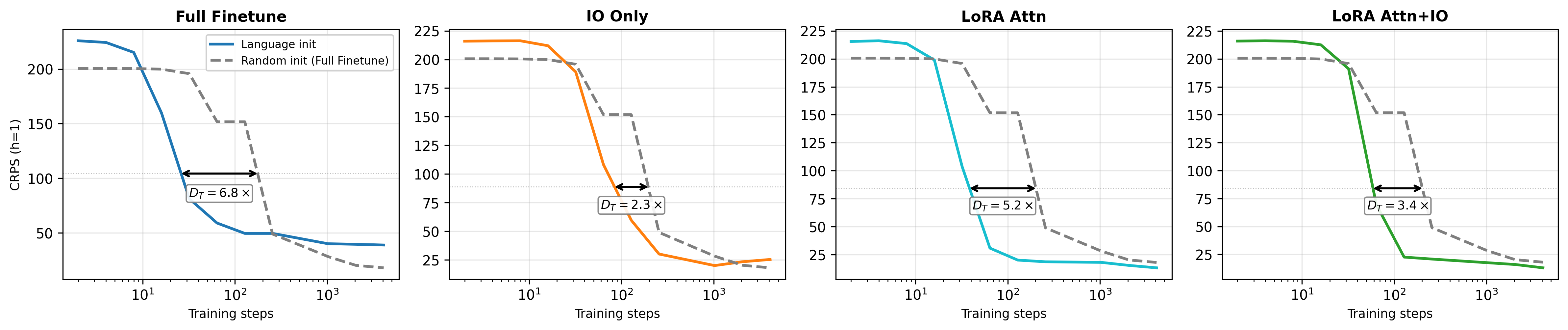}
    \caption{Effective data transfer (see Appendix \ref{app:effective_transfer}) from language pretraining, following the framework of \cite{hernandez2021scalinglawstransfer}. Each panel shows CRPS (h=1) over training steps for a single training regime, comparing language-pretrained initialization (solid) against the full finetuned random initialization (dashed). $D_T$ indicates the multiplicative factor in training steps that the randomly initialized model requires to match the pretrained model at a given performance level.
    }
    \label{fig:data_transfer}
    \vspace{-0.35cm}
\end{figure}

To test whether adaptation from language to time series requires broad changes to pretrained parameters, we compare full finetuning against the parameter-efficient methods from Section \ref{sec:training_regimes}. If low-rank updates recover most of the transfer benefit, then adaptation can be understood as a small perturbation around the pretrained solution rather than a full reconfiguration of the network.

As shown in Figure \ref{fig:data_transfer}, we find that a low-rank adaptation preserves most of the optimization advantage of language pretraining: following the framework of \cite{hernandez2021scalinglawstransfer} (see Appendix \ref{app:effective_transfer}), we define   the {\em effective data transfer} at a given loss level, and see that   full finetuning achieves $D_T = 6.8$, attention-only LoRA achieves $D_T = 5.2$, and the constrained LoRA+IO setting still reaches $D_T = 3.4$, outperforming IO-only tuning ($D_T = 2.3$). Across all settings, pretrained initialization reaches a target CRPS in substantially fewer optimization steps than random initialization, indicating that only a low-dimensional modification of the pretrained weights is needed for time-series forecasting.

\vspace{-0.1in}
\subsection{Effective Rank Dynamics Reveal Selective Compression}
\label{sec:dimensionality}
\vspace{-0.1in}

\textbf{Method. }
To track how representational geometry evolves during training, we compute the effective rank of hidden-state covariance matrices at every saved checkpoint. For each checkpoint and each of the 28 transformer layers, we pass 10{,}000 time-series windows (length 512, drawn from the GiftEval validation split) through the model, collecting hidden states at every position except position~0 (excluded to avoid the attention-sink artifact). For FT and IO-only FT we additionally pass 10{,}000 WikiText-103 sequences to track the residual text representation quality.

At each layer we accumulate the centered covariance matrix $\Sigma = \frac{1}{N}\sum_i \mathbf{h}_i \mathbf{h}_i^\top - \bar{\mathbf{h}}\bar{\mathbf{h}}^\top$ in float64 over all $N$ hidden-state vectors ($N \approx 130{,}000$ per batch), then compute its eigendecomposition. The effective rank is $\mathrm{erank}(\Sigma) = \exp\,\bigl(-\sum_i p_i \log p_i\bigr)$, where $p_i = \lambda_i / \sum_j \lambda_j$ are the normalized eigenvalues~\citep{7098875}. This equals 1 when all variance concentrates on a single direction and equals the ambient dimension $d$ when variance is uniformly spread.

\begin{figure}[!ht]
\centering
\begin{subfigure}[t]{0.48\textwidth}
\centering
\includegraphics[width=\textwidth]{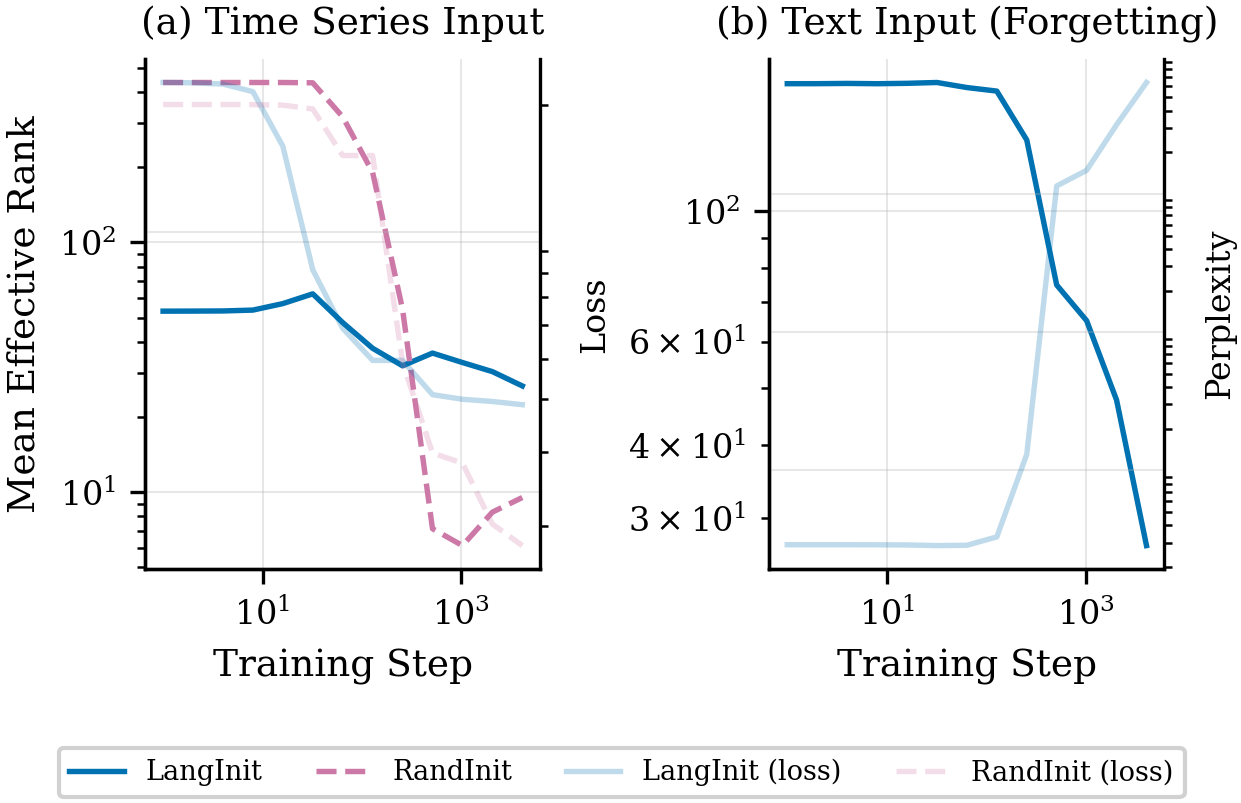}
\end{subfigure}
\hfill
\begin{subfigure}[t]{0.48\textwidth}
\centering
\includegraphics[width=0.9\textwidth]{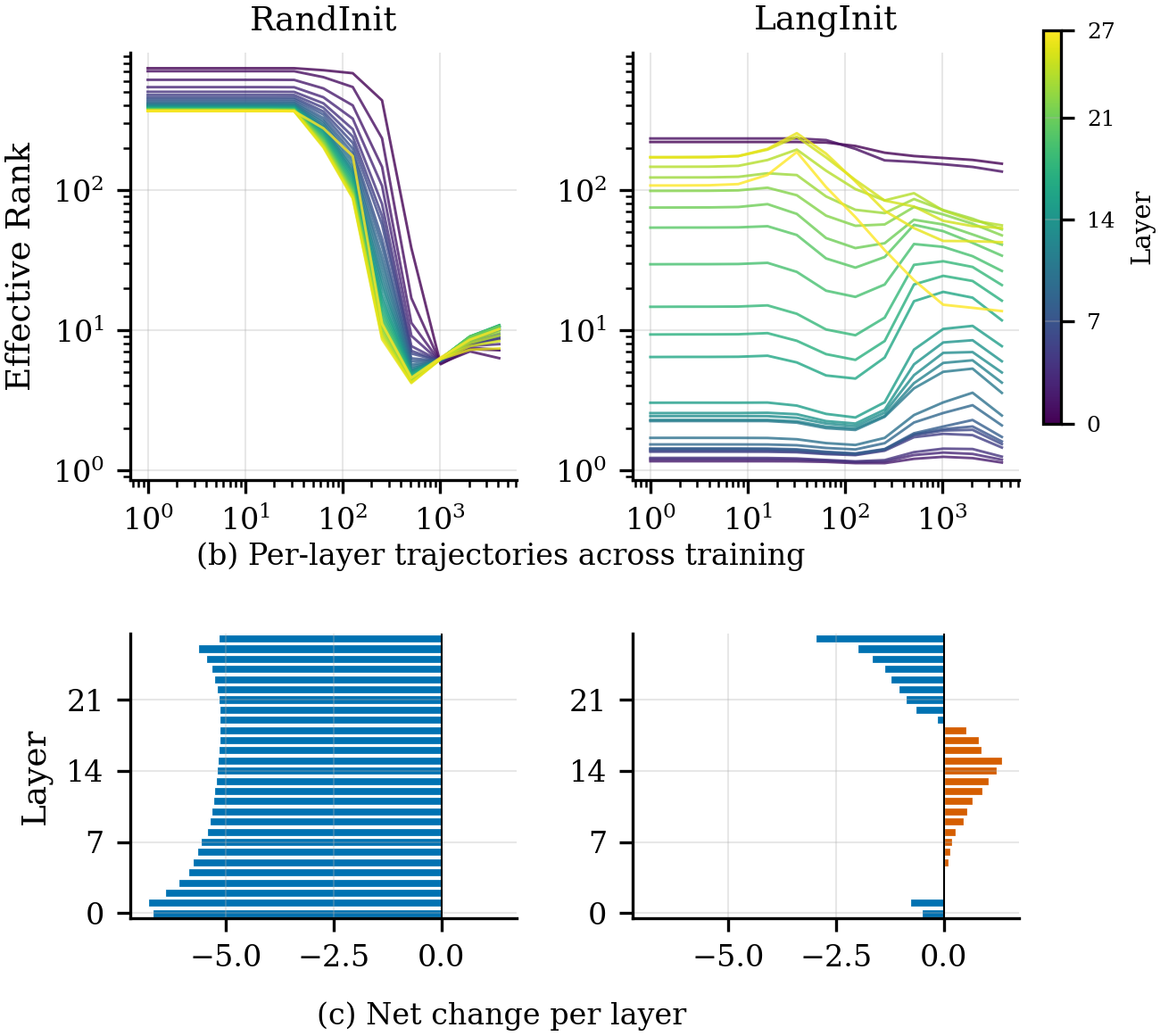}
\end{subfigure}
\caption{\textbf{Effective rank dynamics during training.} \textbf{(a)}~Mean effective rank across all 28 layers. Left: on time-series input, RandInit starts near-isotropic ($\sim$440) and collapses to $\sim$10 within 500 steps; LangInit declines more gradually from the pretrained baseline ($\sim$50) to $\sim$27. Transparent lines show CRPS loss on a secondary axis, where decreasing loss correlates with rank. Right: on text input, LangInit's effective rank drops from $\sim$165 to $\sim$25 (85\% reduction) while perplexity rises, confirming catastrophic forgetting. \textbf{(b)}~Per-layer trajectories (each line = one of 28 layers, colored by depth). RandInit layers collapse as a tight bundle; LangInit middle layers ($\sim$8--17) \emph{increase} their effective rank while early and late layers decrease. \textbf{(c)}~$\log_2$ change (final/initial) per layer. RandInit is uniformly negative while middle layers of LangInit show positive change indicating finetuning redistributes capacity from the network's periphery to its core, reusing abstract language circuitry from middle layers.}
\label{fig:erank_dynamics}
\vspace{-0.3cm}
\end{figure}

\textbf{Results. }
Figure~\ref{fig:erank_dynamics} shows that language-pretrained and randomly initialized models reach time-series representations through different trajectories. RandInit begins with near-isotropic hidden states and rapidly collapses to a low effective rank across nearly all layers, suggesting that it must first compress random representations into a usable low-dimensional structure before specialization can occur. LangInit, in contrast, starts from an already compressed representation and changes more selectively: the mean effective rank decreases only moderately on time-series inputs, while text-input representations collapse sharply, indicating catestrophic forgetting of language-specific directions.

The layerwise dynamics indicate that RandInit undergoes a uniform rank collapse across the network, whereas LangInit redistributes representational capacity: early and late layers lose rank, while middle layers increase effective rank during finetuning. This pattern is consistent with the view that finetuning does not build temporal structure from scratch, but repurposes pretrained middle-layer circuitry and contracts directions less relevant to forecasting. 

The observed dynamics align with the representation phases described in~\citet{li2025tracingrepresentationgeometrylanguage}. The randomly initialized model undergoes the warmup phase marked by rapid representational collapse, followed by an entropy-seeking phase that expands useful directions. The language initialized model seemingly skips the warmup collapse, and move directly toward selective entropy redistribution and compression, preserving variance along directions useful for forecasting while contracting directions less relevant to the new modality.

\vspace{-0.1in}\subsection{Reuse vs.\ Reinvention: FT and RI Find Different Solutions}
\label{sec:reuse_vs_reinvent}
\vspace{-0.1in}

\paragraph{Method.}
To compare how pretrained and randomly initialized models represent simple temporal structure, we pass synthetic waveforms through our models. We visualize hidden-state geometry with 2D PCA, using the middle layer (13) whenever no layer is specified. For training dynamics, we additionally measure phase coherence in the full hidden-state space. Full preprocessing, PCA, and phase-coherence details, along with all-layer visualizations, are provided in Appendix~\ref{app:reuse_reinvention}.

\begin{figure}[!ht]
\centering
\begin{minipage}[t]{0.46\textwidth}
\centering
\includegraphics[width=\textwidth]{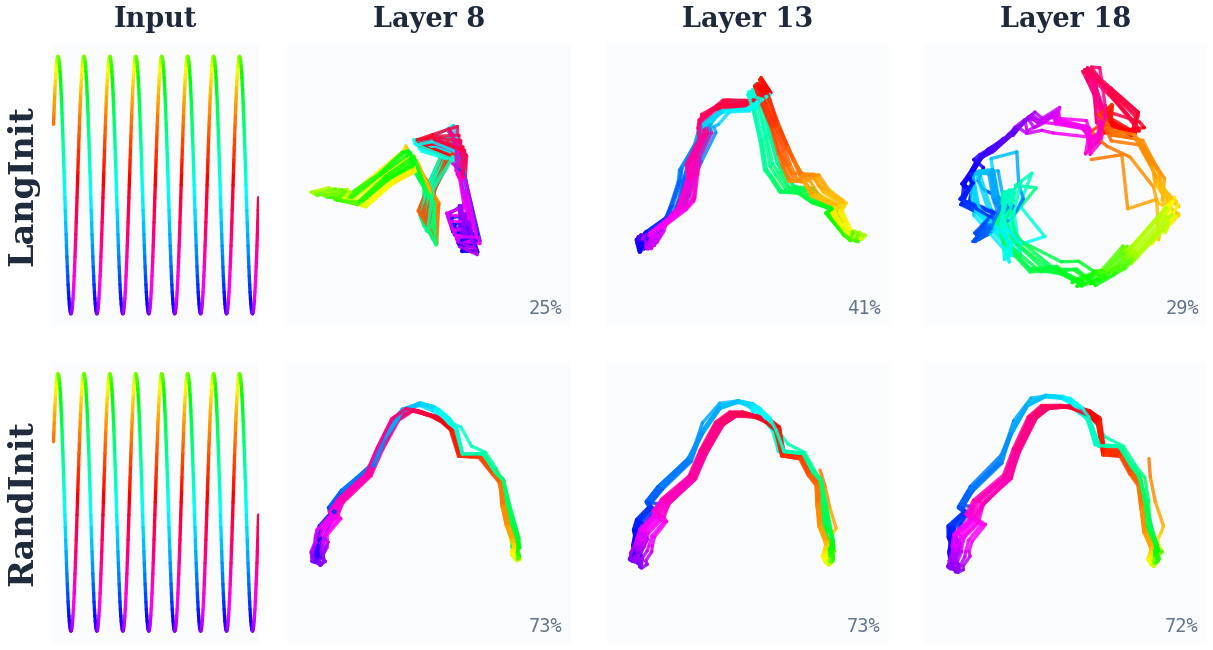}
\end{minipage}\hfill
\begin{minipage}[t]{0.46\textwidth}
\centering
\includegraphics[width=\textwidth]{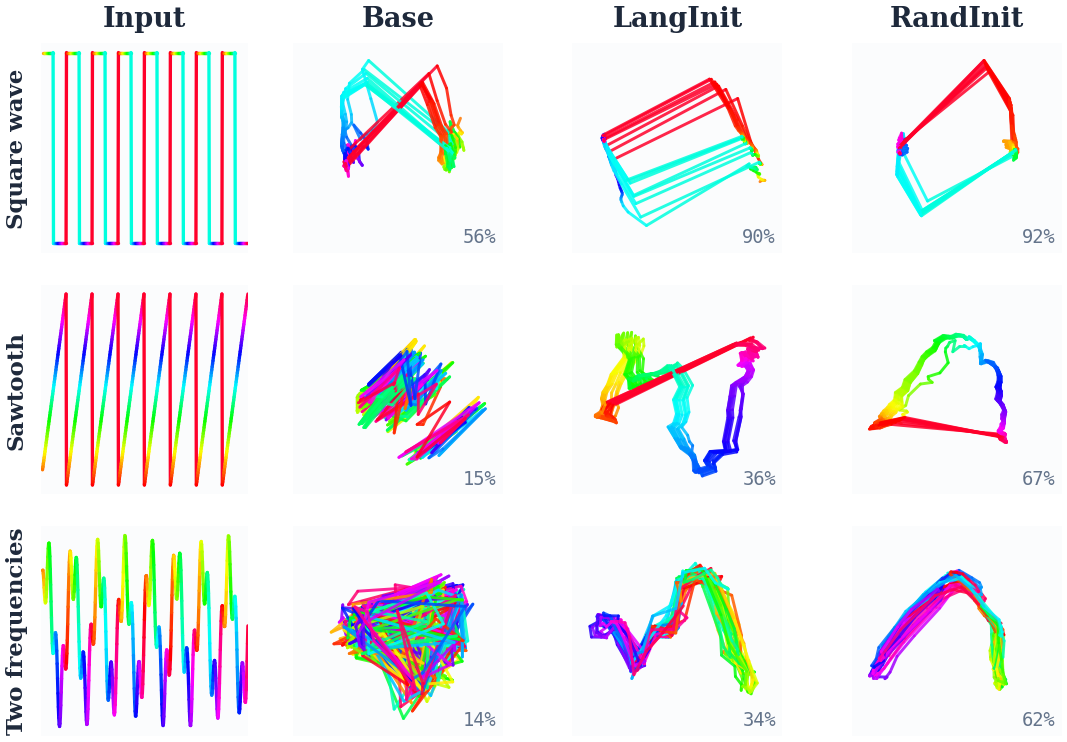}
\end{minipage}
\caption{\textbf{Left: Sine wave representations across layers.} A sine wave (period~64) passed through LangInit (top) and RandInit (bottom) at layers 8, 13, and 18, shown as 2D PCA colored by input phase. RandInit produces clean arcs at every layer (72--73\% PCA variance); LangInit produces complex, layer-varying loops (25--41\%).
\textbf{Right: Hidden-state trajectories for synthetic inputs at Layer~13 (PCA).} Each row is a different input waveform; columns show the pretrained LLM (Base), the finetuned LLM (LangInit), and the randomly-initialized model trained on time series (RandInit). Trajectories are 2D PCA projections of hidden states, colored by input phase. Percentages show variance captured by 2 PCs. Random and Base produce unstructured trajectories. RandInit discovers clean, low-dimensional representations (62--92\% PCA variance) while LangInit creates geometrically complex but structured trajectories (34--98\%) that vary by input type. }
\label{fig:sine_evolution}
\end{figure}
\vspace{-0.1in}
\paragraph{Result.}

Figure~\ref{fig:sine_evolution} (left) shows this contrast for a sine wave across three representative layers. RandInit converges to nearly uniform arcs, with 72--73\% of variance explained by two principal components. LangInit instead produces layer-specific loops with only 25--41\% PCA variance. Thus, both models encode phase, but RandInit represents it in a simple low-dimensional geometry, whereas LangInit distributes the same periodic structure across a richer inherited representation. 
Notice on the (right) plot, the square wave representation from BASE is already sufficiently similar to the learned representation.

Figure~\ref{fig:coherence_evolution} extends the comparison to five synthetic inputs at Layer~13, and across all sine-wave layers in the Appendix (Figures~\ref{fig:evolution_grid_ft}--\ref{fig:evolution_grid_ri}). Here, RandInit learns geometries that are easy to recover with linear projections while LangInit retains more heterogeneous, nonlinear structure inherited from pretraining. 
This is inline with the observation that the eigenspectrum is more spread out in LangInit than RandInit.
The key difference is therefore not whether the models learn temporal structure, but how they learn it. Quantitatively, the trained LangInit and RandInit layer-13 representations are highly aligned across synthetic inputs (CKA $=0.855$--$0.988$), confirming that they recover nearly the same relational geometry up to rotation and scaling (this difference is clear in t-SNE Figure~\ref{fig:synthetic_tsne}).

\subsection{Interpreting the Transferred Features}
\label{sec:crosscoder_features}

\paragraph{Method.}
To test whether the reused directions correspond to interpretable features, we train one sparse crosscoder per layer with a shared encoder and separate pretrained/finetuned decoders. The shared encoder maps pretrained activations on decimal-string time series and finetuned activations on binned time series into a common 4{,}096-dimensional Top-$K$ latent space; features are ranked by cross-domain firing between time series and WikiText.

\vspace{-0.1in}
\begin{table}[h]
\centering
\small
\caption{Examples of shared PT--FT crosscoder features. Shared features concentrate in layers~7--10; the examples below link concrete time-series motifs to coherent WikiText genres. (See Appendix~\ref{app:crosscoder})}
\label{tab:crosscoder_features_main}
\begin{tabularx}{\linewidth}{clXX}
\toprule
Layer & Feature & Time-series motif & WikiText motif \\
\midrule
10 & 1712 & magnitude jumps / plateaus & measurements and unit conversions \\
9  & 2469 & volatile regime changes & tropical cyclone narratives \\
7  & 3888 & isolated spikes & timestamped naval battles \\
8  & 2567 & missing / NaN windows & \texttt{<unk>} and incomplete references \\
\bottomrule
\end{tabularx}
\end{table}
A few top activated features are shown in Table \ref{tab:crosscoder_features_main}. These features suggest that the aligned LangInit subspace is populated by reusable temporal primitives that may correlate to text semantics, not just arbitrary low-rank directions. Appendix~\ref{app:circuit} provides preliminary causal evidence for the same picture: zero-ablation identifies a superadditive Layer~1 circuit (head~L1H4 $\leftrightarrow$ MLP\textsubscript{L1}) that is critical for periodic time-series prediction in IO-only FT and most affects pretrained text with sequential repetitive structure. 

\vspace{-0.1in}
\section{Limitations}
\label{sec:limitations} \vspace{-0.1in}

All experiments use a single backbone (Qwen3-0.6B), one training corpus (GiftEval), and one tokenization scheme (1024 uniform bins); the geometric account has not been verified at larger scale or on alternative architectures. The linear probe establishes that the pretrained manifold is \emph{compatible} with real time series, not that it contains the optimal forecasting subspace. The crosscoder analysis surfaces shared primitives in middle layers but does not exhaustively characterize the transferred subspace, and the causal circuit evidence (Appendix) is preliminary. We do not fully isolate whether gradient coherence (Sec.~\ref{sec:curvature}) is attributable to manifold geometry versus initialization statistics. Finally, ``geometric rather than semantic'' describes the dominant mechanism we observe.

\vspace{-0.1in}
\section{Discussion and Conclusion}
\vspace{-0.1in}

We investigated why language-pretrained transformers transfer to time-series forecasting and found that the mechanism is geometric rather than semantic. Pretrained representations already contain directions aligned with realistic temporal trajectories, enabling optimization to descend immediately through coherent gradients and favorable curvature. Finetuning then acts primarily as low-rank specialization over a pretrained sequential manifold rather than learning forecasting structure from scratch. The reused subspace captures structural primitives shared across modalities, including periodicity, regime changes, magnitude shifts, and missingness. In contrast, randomly initialized models eventually recover competitive forecasting performance only after constructing a new and geometrically simpler representation space.
Our results suggest that this mechanism extends beyond language models: any sufficiently rich autoregressive sequence model with latent temporal structure may support similar transfer. This perspective reconciles prior conflicting findings by explaining why transfer is strongest in low-data and distribution-shift regimes, where pretrained directions provide structure that limited forecasting data alone cannot establish. It also provides a principled explanation for the effectiveness of LoRA and related parameter-efficient methods, since if transfer is fundamentally direction selection, low-rank adaptation is the natural mechanism for exploiting it. 

\section{Acknowledgment}
We thank \href{https://www.42.com}{42.com} for providing computational resources that supported this work. We also acknowledge the \href{https://www.amd.com/en/corporate/university-program/ai-hpc-cluster.html}{AMD University Program AI \& HPC Cluster} for additional compute resources used in this research.

\small
\bibliographystyle{plainnat}
\bibliography{references}

\newpage
\appendix
\section{Experimental Setup}
\label{sec:appendix_experimental_setup}

\subsection{Hyperparameters and Reproducibility}
\label{sec:hyperparams}
\begin{table}[h]
\centering
\caption{Complete hyperparameter configuration for all experiments. All runs share the same configuration except where noted.}
\label{tab:hyperparameters}
\begin{tabular}{lll}
\toprule
\textbf{Category} & \textbf{Parameter} & \textbf{Value} \\
\midrule
\multirow{2}{*}{Model}
& Base model          & Qwen3-0.6B \\
& Architecture        & Qwen3ForCausalLM and Qwen defaults \\
\midrule
\multirow{7}{*}{Tokenizer}
& Scaling             & Z-score (mean/std from context) \\
& Binning             & Uniform \\
& Vocab size ($V$)    & 1024 \\
& Bin range            & $[-5, 5]$ \\
& Use EOS token       & No \\
\midrule
\multirow{6}{*}{Training}
& Optimizer           & AdamW \\
& Learning rate       & $1 \times 10^{-4}$, $3 \times 10^{-4}$, $1 \times 10^{-3}$, or $3 \times 10^{-3}$ \\
& LR schedule         & Linear warmup + cosine decay \\
& Warmup ratio        & 3\% \\
& LR end factor       & $1 \times 10^{-4}$ \\
& Precision           & bf16 (mixed) \\
\midrule
\multirow{2}{*}{Batching}
& Effective batch size & 128 \\
& Epoch length        & 5{,}000 steps \\
\midrule
\multirow{4}{*}{Data}
& Dataset             & GiftEval Pretrain (152 sub-datasets) \\
& Context length      & 512 \\
& Target length       & 64 \\
& Window stride & 600 \\
\midrule
\multirow{3}{*}{Loss}
& Loss function       & Quantile (pinball) loss \\
& Quantiles           & $\{0.1, 0.2, \ldots, 0.9\}$ \\
& Softmax temperature & $10^{-2}$ \\
\midrule
\multirow{4}{*}{Evaluation}
& Eval samples        & 1{,}000 extracted from the dataset (see script) \\
& Eval horizons       & $h \in \{1, 64\}$ \\
& Eval strategy       & Autoregressive \\
\midrule
\multirow{5}{*}{LoRA}
& Rank ($r$)          & 4, 8 \\
& Alpha ($\alpha$)    & 16, 32 \\
& Dropout             & 0.05 \\
& Bias                & None \\
& Targets (Attn)      & \texttt{q\_proj, k\_proj, v\_proj, o\_proj} \\
\midrule
\multirow{3}{*}{Reproducibility}
& Random seed         & 420 \\
& Seeded modules      & Python, NumPy, PyTorch (CPU + CUDA) \\
& Compute Usage       & About 12 hours on 8 Nvidia A100 per model \\
\bottomrule
\end{tabular}
\end{table}

\subsection{Evaluation Metrics}
\label{sec:appendix_metrics}

This section details the evaluation metrics implemented for assessing the forecasting performance of the models. For all fixed-point metrics, the 0.5 quantile (median) is extracted from the probabilistic forecasts and used as the deterministic prediction.

\textbf{Aggregation Strategy and Implementation:} In our evaluation methodology, we follow closely the GluonTS library \citep{alexandrov2020gluonts}, particularly, the aggregation approach.
All metrics described below are computed \textit{per-sequence first}, and then averaged across all sequences in the dataset. Specifically, the error or score is aggregated over the forecast horizon $H$ for each individual time series, and the final reported metric is the unweighted mean of these per-sequence scores. This macro-averaging ensures that every sequence contributes equally to the final evaluation, regardless of its absolute magnitude or scale. 

To ensure robust evaluation, any time steps containing \text{NaN} values in the ground truth are dynamically excluded from computation by masking. Additionally, a small utility constant ($\epsilon = \text{1e-8}$) is added to denominators in percentage and scale-normalized metrics to prevent division by zero.

\subsubsection{Point Forecast Metrics}

\paragraph{Basic Error Metrics}
Mean Squared Error (MSE) quantifies prediction accuracy by heavily penalizing larger errors, making it sensitive to outliers. Root Mean Squared Error (RMSE) provides the error in the original data units while retaining MSE's sensitivity.
$$\text{MSE} = \frac{1}{H} \sum_{t=1}^{H} (y_t - \hat{y}_t)^2$$
$$\text{RMSE} = \sqrt{\text{MSE}}$$
where $H$ is the forecast horizon, $y_t$ is the true value, and $\hat{y}_t$ is the predicted median.

Mean Absolute Error (MAE) applies a linear penalty to measure the average absolute difference:
$$\text{MAE} = \frac{1}{H} \sum_{t=1}^{H} |y_t - \hat{y}_t|$$

\paragraph{Normalized Metrics}
Mean Absolute Scaled Error (MASE) measures forecast accuracy relative to a naive baseline, allowing for cross-series comparison across data with vastly different scales.
$$\text{MASE} = \frac{\text{MAE}}{\frac{1}{H-m} \sum_{t=m+1}^{H} |y_t - y_{t-m}|}$$
The seasonality parameter $m$ is selected based on the expected periodic patterns of the timeframe ($m=1$ for a general naive baseline, $m=60$ for 1-minute data, and $m=24$ for 1-hour data).

To further achieve scale-invariance, RMSE and absolute deviations are normalized by the scale of the data (mean or sum of absolute values) to compute the Normalized Root Mean Squared Error (NRMSE) and Normalized Deviation (ND):
$$\text{NRMSE} = \frac{\sqrt{\frac{1}{H} \sum_{t=1}^{H} (y_t - \hat{y}_t)^2}}{\frac{1}{H} \sum_{t=1}^{H} |y_t|}$$
$$\text{ND} = \frac{\sum_{t=1}^{H} |y_t - \hat{y}_t|}{\sum_{t=1}^{H} |y_t|}$$

\paragraph{Percentage-Based Metrics}
Mean Absolute Percentage Error (MAPE) provides a scale-independent metric as a percentage. To address MAPE's asymmetry problem and instability near zero, the Symmetric Mean Absolute Percentage Error (sMAPE) equally penalizes over-predictions and under-predictions:
$$\text{MAPE} = \frac{100}{H} \sum_{t=1}^{H} \frac{|y_t - \hat{y}_t|}{|y_t| + \epsilon}$$
$$\text{sMAPE} = \frac{200}{H} \sum_{t=1}^{H} \frac{|y_t - \hat{y}_t|}{|y_t| + |\hat{y}_t| + \epsilon}$$

\subsubsection{Probabilistic Metrics}

\paragraph{Continuous Ranked Probability Score (CRPS)}
The approximated CRPS measures the accuracy of probabilistic forecasts by evaluating the entire predictive distribution using quantile loss ($\text{QL}$).
$$\text{CRPS}_{\text{approx}} = 2 \sum_{q} w_q \cdot \text{QL}_q(y_{\text{true}}, y_{\text{pred}})$$
where $\text{QL}_q(y, \hat{y}) = (q - \mathbf{1}_{y \leq \hat{y}}) \cdot (y - \hat{y})$ and $w_q$ represents the discrete weight calculated based on the distance between quantile levels.

\paragraph{Weighted Mean Absolute Percentage Quantile Loss (wMAPE)}
To compare probabilistic performance across heterogeneous time series, we implement a scale-invariant version of quantile loss weighted by the sum of absolute true values:
$$\text{wMAPE} = \frac{1}{Q} \sum_{q=1}^{Q} \frac{\sum_{t=1}^{H} \text{QL}_q(y_t, \hat{y}_{q,t})}{\sum_{t=1}^{H} |y_t|}$$

\subsubsection{Directional and Correlation Metrics}

\paragraph{Directional Accuracy (DA)}
DA evaluates how often the model correctly predicts the direction of movement relative to a historical anchor point $y_0$ (the last observed value). 
$$\text{DA} = \frac{1}{H} \sum_{t=1}^{H} \mathbf{1}_{\text{sign}_{\tau}(\hat{y}_t - y_0) = \text{sign}_{\tau}(y_t - y_0)}$$

\paragraph{Correlation Metrics}
To evaluate the model's ability to capture trend dynamics independently of absolute magnitude errors, we utilize Pearson correlation coefficients. Pearson evaluates the linear relationship between predictions and ground truth.

\subsection{Experimental Results}
\begin{figure}
    \centering
    \includegraphics[width=\linewidth]{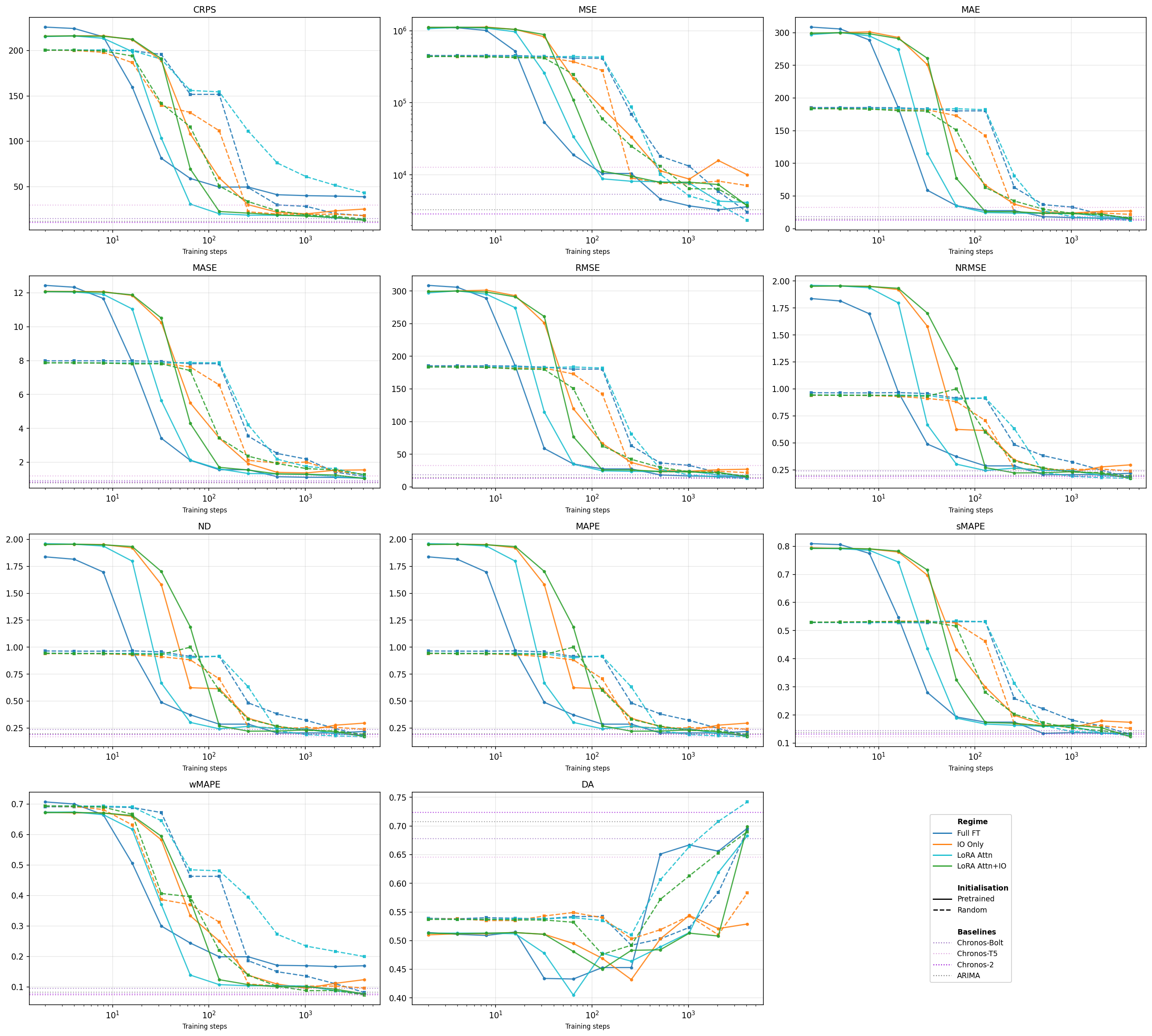}
    \caption{Training progression of all $h=1$ forecasting metrics across training steps for four training regimes. Solid lines denote language-pretrained initialization (Qwen3-0.6B); dashed lines denote random initialization. Horizontal dotted lines indicate baseline performance (Chronos-T5, Chronos-Bolt, Chronos-2, ARIMA). Language-initialized models consistently converge earlier than their randomly initialized counterparts across all metrics, with the gap most pronounced in the first 100 steps. Both initializations converge in performance by 4096 steps, reaching levels competitive with established forecasting baselines.}
    \label{fig:training_progression_h1}
\end{figure}

Here we provide the complete set of evaluation results across all training regimes and metrics. Figure~\ref{fig:training_progression_h1} shows the training progression of all eleven h=1 metrics over 4096 training steps, revealing a consistent pattern across metrics: language-pretrained models begin converging within the first 10–100 steps, while randomly initialized models require half an order of magnitude more training to reach comparable performance. By step 4096, both initialization strategies converge to similar error levels across all regimes, confirming that the pretrained weights accelerate optimization rather than alter the final solution. Tables~\ref{tab:h1-metrics} and~\ref{tab:h64-metrics} report the full numerical results at step 128 for single-step (h=1) and multi-step (h=64) forecasting respectively, alongside four established baselines and a zero-shot Qwen3 text baseline. At this early checkpoint, the transfer advantage of language pretraining is clearly visible: pretrained LoRA Attn achieves a CRPS of 20.10 compared to 154.5 for its randomly initialized counterpart, while all pretrained regimes outperform the random initializations that have not yet begun to converge. The Qwen3 text baseline performs orders of magnitude worse than all trained models, confirming that the language model's pretrained representations require domain-specific finetuning to be useful for time series forecasting.

\subsection{Effective Transfer}\label{app:effective_transfer}

Let $\mathcal L_R(d)$ and $\mathcal L_P(d)$ denote the validation losses achieved after training for $d$ time series tokens using random and pre-trained initializations, respectively.
For a given validation loss $\ell$, $\mathcal L_\cdot^{-1}(\ell)$ therefore denotes the amount of data required to train a model to reach the loss level $\ell$ begining from a given initialization.
We define transfer as ``effective'' or ``positive'' when starting the timeseries model training with the language model's weights allows us to achieve the same validation loss with less data than a model initialised randomly. This data quantity difference is what we refer to as the amount of data we ``saved''. Concretely, the \textit{effective data transferred} $D_T(\ell)$ for a target loss $\ell$ is defined as the difference of these data amounts:
\[
    D_T(\ell) := \mathcal L_R^{-1}(\ell) - L_P^{-1}(\ell).
\]
A positive $D_T(\ell)$ indicates that initializing the model from pre-trained language weights requires fewer training examples to reach the loss $\ell$ compared to random initialization, signifying an effective transfer from the upstream task to time series forecasting.

\vspace{12em}

\begin{table}[h]
\centering
\caption{Single-step ($h{=}1$) forecasting performance on the held-out evaluation set. All NanoTS variants use Qwen3-0.6B at training step 128. Best value per metric within each section is \textbf{bolded}. $\uparrow$\! : higher is better; $\downarrow$\! : lower is better.}
\label{tab:h1-metrics}
\resizebox{\textwidth}{!}{\begin{tabular}{ll ccccccccccc}
\toprule
& \textbf{Model} & \textbf{CRPS}$\downarrow$ & \textbf{MSE}$\downarrow$ & \textbf{MAE}$\downarrow$ & \textbf{MASE}$\downarrow$ & \textbf{RMSE}$\downarrow$ & \textbf{NRMSE}$\downarrow$ & \textbf{ND}$\downarrow$ & \textbf{MAPE}$\downarrow$ & \textbf{sMAPE}$\downarrow$ & \textbf{wMAPE}$\downarrow$ & \textbf{DA}$\uparrow$ \\
\midrule
\multirow{4}{*}{\rotatebox[origin=c]{90}{\scriptsize Pretrained}}
& Full Finetune     & 49.51 & 10461 & 27.60 & \textbf{1.561} & 27.60 & 0.286 & 0.286 & 0.286 & 0.175 & 0.199 & 0.453 \\
& IO Only            & 59.56 & 84846 & 66.53 & 3.430 & 66.53 & 0.614 & 0.614 & 0.614 & 0.300 & 0.250 & 0.469 \\
& LoRA Attn          & \textbf{20.10} & \textbf{8818} & \textbf{24.68} & 1.607 & \textbf{24.68} & \textbf{0.243} & \textbf{0.243} & \textbf{0.243} & \textbf{0.168} & \textbf{0.107} & \textbf{0.478} \\
& LoRA Attn+IO       & 22.54 & 11220 & 26.11 & 1.702 & 26.11 & 0.270 & 0.270 & 0.270 & 0.174 & 0.124 & 0.450 \\
\midrule
\multirow{4}{*}{\rotatebox[origin=c]{90}{\scriptsize Random Init}}
& Full Finetune     & 151.8 & 415339 & 180.4 & 7.819 & 180.4 & 0.914 & 0.914 & 0.914 & 0.532 & 0.463 & \textbf{0.542} \\
& IO Only            & 111.6 & 280932 & 142.3 & 6.565 & 142.3 & 0.706 & 0.706 & 0.706 & 0.462 & 0.313 & 0.540 \\
& LoRA Attn          & 154.5 & 430387 & 182.2 & 7.872 & 182.2 & 0.917 & 0.917 & 0.917 & 0.532 & 0.481 & 0.535 \\
& LoRA Attn+IO       & \textbf{50.68} & \textbf{59864} & \textbf{62.90} & \textbf{3.438} & \textbf{62.90} & \textbf{0.599} & \textbf{0.599} & \textbf{0.599} & \textbf{0.281} & \textbf{0.220} & 0.476 \\
\midrule
\multirow{5}{*}{\rotatebox[origin=c]{90}{\scriptsize Baselines}}
& Chronos-Bolt       & 14.94 & 5409 & 18.91 & 0.934 & 18.91 & 0.240 & 0.240 & 0.240 & 0.146 & 0.096 & 0.678 \\
& Chronos-T5         & 29.88 & 12764 & 32.99 & 1.212 & 32.99 & \textbf{0.173} & \textbf{0.173} & \textbf{0.173} & \textbf{0.124} & \textbf{0.073} & 0.646 \\
& Chronos-2          & \textbf{10.89} & \textbf{2886} & \textbf{13.38} & 0.830 & \textbf{13.38} & 0.192 & 0.192 & 0.192 & 0.133 & 0.077 & \textbf{0.724} \\
& ARIMA              & 11.94 & 3327 & 14.86 & \textbf{0.805} & 14.86 & 0.201 & 0.201 & 0.201 & 0.138 & 0.083 & 0.708 \\
& Qwen3 Text         & 213.5 & 2.6M & 213.5 & 376667 & 213.5 & 39508 & 39508 & 39508 & 0.563 & 19754 & 0.503 \\
\bottomrule
\end{tabular}
}
\end{table}

\begin{figure}
    \centering
    \includegraphics[width=\linewidth]{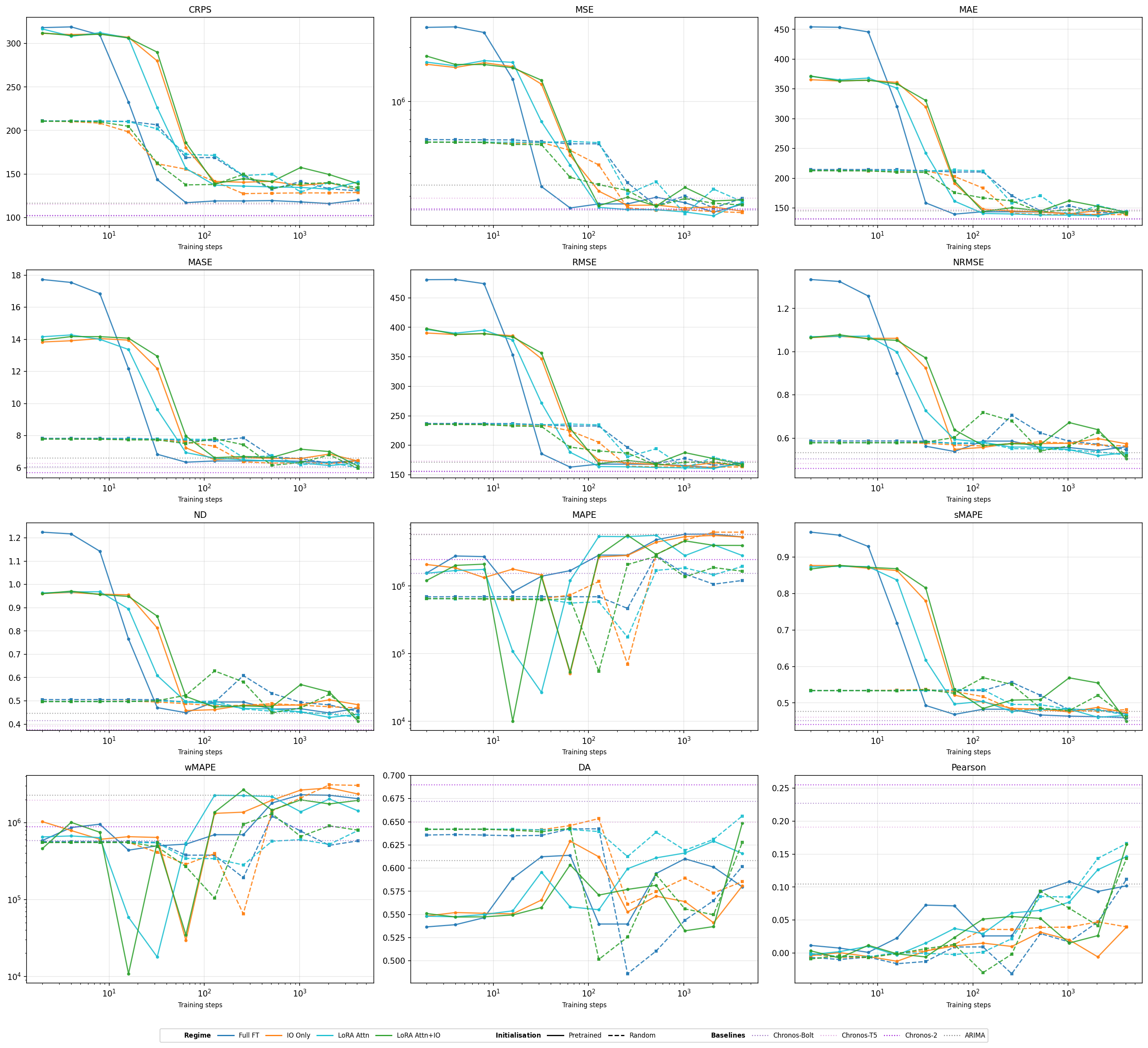}
    \caption{Training progression of all $h=64$ forecasting metrics across training steps for four training regimes. Solid lines denote language-pretrained initialization (Qwen3-0.6B); dashed lines denote random initialization. Horizontal dotted lines indicate baseline performance (Chronos-T5, Chronos-Bolt, Chronos-2, ARIMA). Language-initialized models consistently converge earlier than their randomly initialized counterparts across all metrics, with the gap most pronounced in the first 100 steps. Both initializations converge in performance by 4096 steps, reaching levels competitive with established forecasting baselines.}
    \label{fig:training_progression_h64}
\end{figure}

\begin{table}[t]
\centering
\caption{Multi-step ($h{=}64$) forecasting performance on the held-out evaluation set. All NanoTS variants use Qwen3-0.6B at training step 128. Best value per metric within each section is \textbf{bolded}. $\uparrow$\!: higher is better; $\downarrow$\!: lower is better.}
\label{tab:h64-metrics}
\resizebox{\textwidth}{!}{\begin{tabular}{ll ccccccccccccc}
\toprule
& \textbf{Model} & \textbf{CRPS}$\downarrow$ & \textbf{MSE}$\downarrow$ & \textbf{MAE}$\downarrow$ & \textbf{MASE}$\downarrow$ & \textbf{RMSE}$\downarrow$ & \textbf{NRMSE}$\downarrow$ & \textbf{ND}$\downarrow$ & \textbf{MAPE}$\downarrow$ & \textbf{sMAPE}$\downarrow$ & \textbf{wMAPE}$\downarrow$ & \textbf{DA}$\uparrow$ & \textbf{Pearson}$\uparrow$ \\
\midrule
\multirow{4}{*}{\rotatebox[origin=c]{90}{\scriptsize Pretrained}}
& Full Finetune     & \textbf{119.0} & 271341 & 143.7 & \textbf{6.424} & 168.2 & 0.586 & 0.495 & \textbf{2.85M} & \textbf{0.483} & \textbf{696208} & 0.540 & 0.026 \\
& IO Only            & 141.5 & 319226 & 147.9 & 6.502 & 174.9 & \textbf{0.557} & \textbf{0.462} & 2.68M & 0.503 & 1.32M & \textbf{0.612} & 0.015 \\
& LoRA Attn          & 137.2 & \textbf{259652} & \textbf{140.8} & 6.602 & \textbf{164.0} & 0.581 & 0.487 & 5.39M & 0.505 & 2.27M & 0.555 & 0.029 \\
& LoRA Attn+IO       & 138.8 & 265546 & 144.0 & 6.628 & 169.0 & 0.566 & 0.473 & 2.82M & 0.485 & 1.36M & 0.571 & \textbf{0.052} \\
\midrule
\multirow{4}{*}{\rotatebox[origin=c]{90}{\scriptsize Random Init}}
& Full Finetune     & 168.9 & 583696 & 210.5 & 7.700 & 232.7 & 0.575 & 0.494 & 693452 & 0.534 & 377358 & 0.642 & 0.009 \\
& IO Only            & 141.3 & 445540 & 183.5 & \textbf{7.341} & 205.0 & \textbf{0.563} & \textbf{0.478} & 1.17M & \textbf{0.517} & 394850 & \textbf{0.653} & \textbf{0.036} \\
& LoRA Attn          & 171.4 & 592435 & 212.3 & 7.770 & 234.7 & 0.579 & 0.499 & 583786 & 0.536 & 339884 & 0.639 & 0.001 \\
& LoRA Attn+IO       & \textbf{137.9} & \textbf{346961} & \textbf{166.8} & 7.812 & \textbf{190.5} & 0.719 & 0.628 & \textbf{55012} & 0.569 & \textbf{104026} & 0.502 & $-$0.030 \\
\midrule
\multirow{5}{*}{\rotatebox[origin=c]{90}{\scriptsize Baselines}}
& Chronos-Bolt       & \textbf{101.9} & \textbf{251450} & \textbf{131.7} & 6.068 & 156.8 & 0.504 & 0.416 & \textbf{1.55M} & 0.452 & \textbf{587680} & 0.672 & 0.227 \\
& Chronos-T5         & 115.2 & 292023 & 147.4 & 6.291 & 173.7 & 0.484 & 0.393 & 5.84M & 0.462 & 1.98M & 0.650 & 0.191 \\
& Chronos-2          & 102.4 & 255127 & 132.0 & \textbf{5.709} & \textbf{156.2} & \textbf{0.460} & \textbf{0.376} & 2.48M & \textbf{0.441} & 888025 & \textbf{0.690} & \textbf{0.255} \\
& ARIMA              & 116.6 & 345041 & 145.4 & 6.609 & 171.8 & 0.534 & 0.446 & 5.80M & 0.478 & 2.27M & 0.608 & 0.105 \\
& Qwen3 Text         & 489.1 & 213.1M & 489.1 & 9911.0 & 720.8 & 5672.0 & 824.6 & 3.23M & 0.568 & 1.62M & 0.589 & $-$0.048 \\
\bottomrule
\end{tabular}}
\end{table}

 \clearpage

\section{Linear Mapping Experiment}

\begin{table}[H]
\centering
\caption{Fair top-$K$ comparison across ablation conditions. For each condition, we take the best-$K$ unique matches (deduplicated) and report mean nearest-neighbor MSE. This controls for diversity: conditions with fewer unique matches are only compared at $K$ values they can support. ``---'' indicates fewer than $K$ unique matches available. Text + PT achieves the lowest MSE at every $K$.}
\label{app:fair_comparison}
\begin{tabular}{rcccc}
\toprule
$K$ & text + PT & text + RandInit & rand + PT & rand + RandInit \\
\midrule
4   & \textbf{0.254} & 0.383 & 0.330 & 0.412 \\
54  & \textbf{0.350} & 0.721 & ---   & 0.755 \\
99  & \textbf{0.383} & 0.825 & ---   & 0.862 \\
200 & \textbf{0.441} & 0.971 & ---   & 1.004 \\
439 & \textbf{0.535} & ---   & ---   & ---   \\
686 & \textbf{0.639} & ---   & ---   & ---   \\
\bottomrule
\end{tabular}
\end{table}

\section{Reuse vs.\ Reinvention: Additional Figures}
\label{app:reuse_reinvention}

\paragraph{Method details.}
We generate five length-512 synthetic inputs: a sine wave (period~64), a square wave (period~64), a sawtooth wave (period~64), a two-frequency signal $\sin(2\pi t/64) + 0.5\sin(2\pi t/17)$, and a linear trend with oscillation $t/T + 0.3\sin(2\pi t/80)$. Each input is independently $z$-score normalized, clipped to $[-5,5]$, and uniformly binned into 1024 tokens using the same discretization as in the forecasting experiments. After passing the tokenized sequence through each model, we extract hidden states and fit PCA independently for each model--input pair; therefore, the plots compare within-trajectory structure and variance explained rather than absolute PCA axes across models. We exclude the first five positions to reduce attention-sink effects and color points by position modulo the dominant period of the input.

For the phase-coherence analysis, we compute Euclidean distances $d_{ij}$ between hidden states at positions $i$ and $j$ in the full 1024-dimensional hidden-state space, again excluding the first five positions. For an input with period $P$, positions are treated as same-phase when $i \bmod P = j \bmod P$. We define
\begin{equation}
\mathrm{coherence}
= \frac{\mathrm{mean}\!\left(d_{ij} \mid i \bmod P = j \bmod P\right)}
{\mathrm{mean}\!\left(d_{ij} \mid \text{all pairs}\right)}.
\end{equation}
If the model encodes periodic structure, same-phase hidden states should be close relative to arbitrary pairs, so lower values indicate stronger phase structure. In the training-dynamics figure we plot $1-\mathrm{coherence}$ so that higher values correspond to stronger periodic encoding.

\begin{figure}[!ht]
\centering
\includegraphics[width=0.85\textwidth]{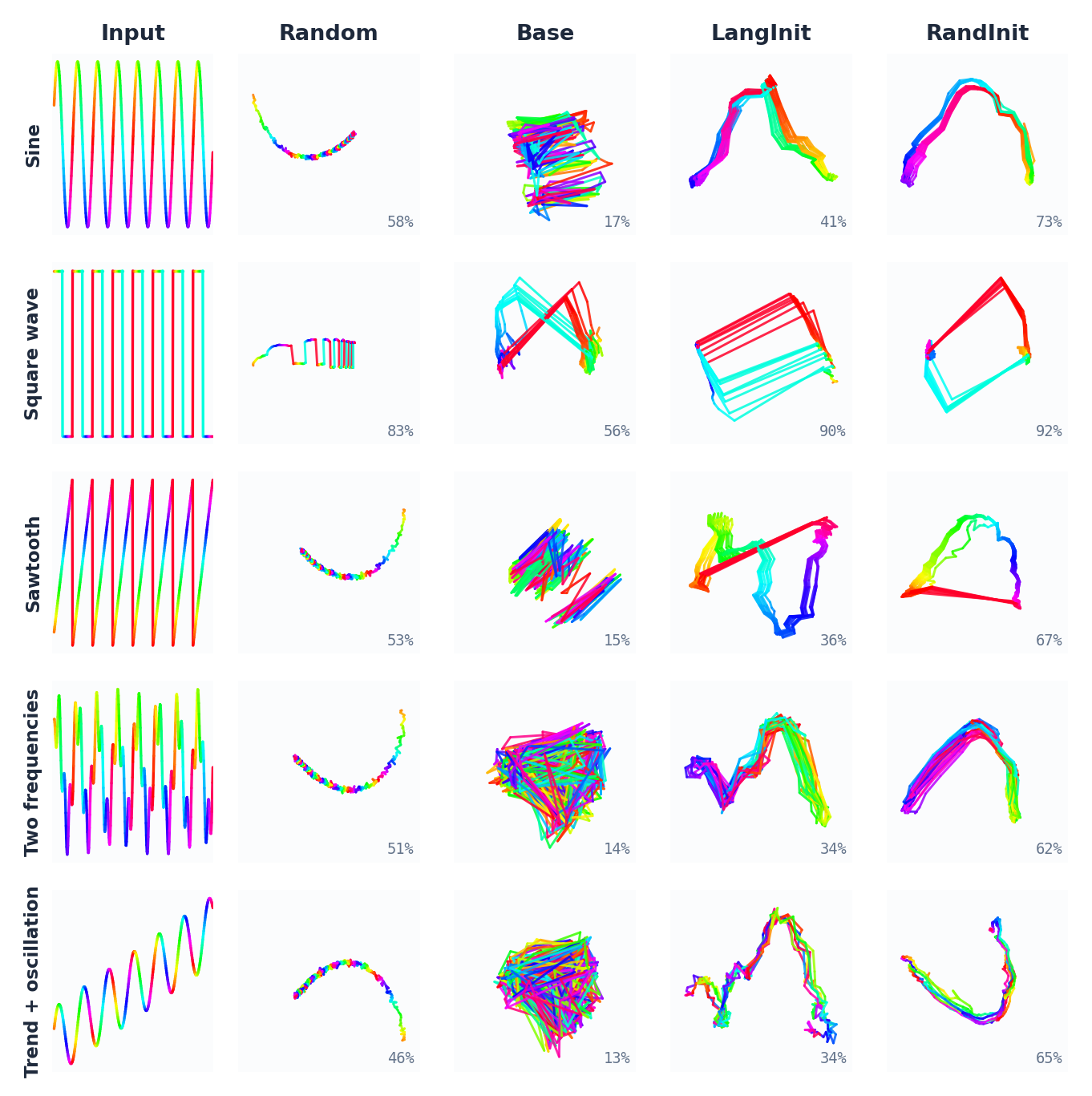}
\caption{\textbf{Hidden-state trajectories for synthetic inputs at Layer~13 (2 PCA).} Trajectories are 2D PCA projections of hidden states, colored by input phase. Percentages show variance captured by 2 PCs. Random and Base produce unstructured trajectories. RandInit discovers clean, low-dimensional representations (62--92\% PCA variance) while LangInit creates geometrically complex but structured trajectories (34--98\%) that vary by input type. }
\label{fig:synthetic_pca}
\end{figure}

\begin{figure}[!ht]
\centering
\includegraphics[width=0.85\textwidth]{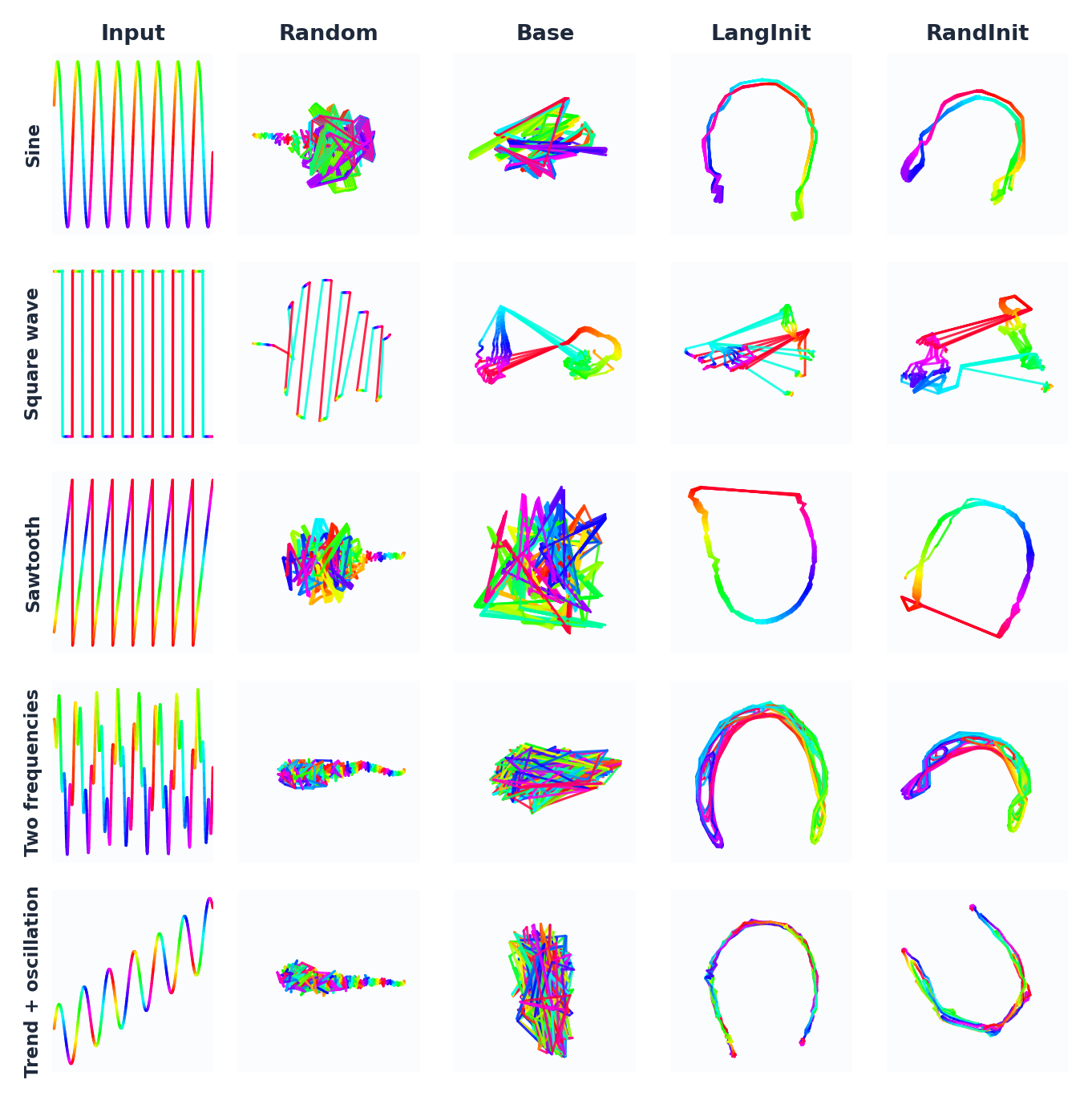}
\caption{\textbf{Hidden-state trajectories for synthetic inputs at Layer~13 (t-SNE).} Same setup as Figure~\ref{fig:synthetic_pca} but using t-SNE (perplexity~30) instead of PCA. Despite the different global geometries revealed by PCA, t-SNE shows that LangInit and RandInit develop similar local neighborhood structure: periodic inputs form smooth, phase-ordered curves in both models, confirming that both arrive at functionally equivalent representations through different geometric paths.}
\label{fig:synthetic_tsne}
\end{figure}

\begin{figure}[!ht]
\centering
\includegraphics[width=\textwidth]{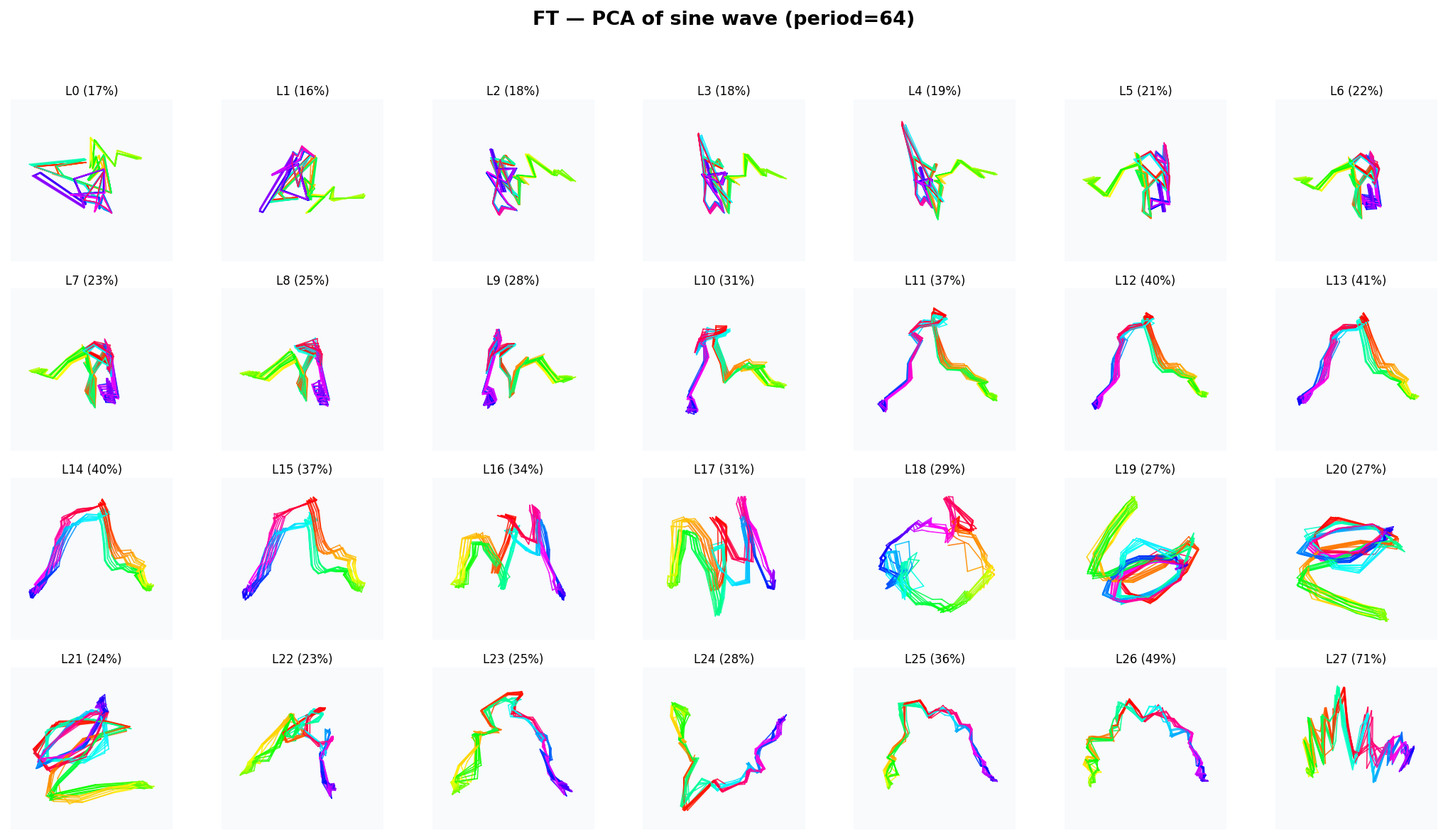}
\caption{\textbf{LangInit: sine wave PCA trajectories across all 28 layers.} Each subplot shows the 2D PCA projection of hidden states from a sine wave (period~64) at one transformer layer, colored by input phase. Percentages show variance explained by 2 PCs. LangInit exhibits layer-specific geometry with varying complexity and moderate PCA variance (17--49\%), reflecting the rich, heterogeneous representations inherited from language pretraining.}
\label{fig:pca_grid_ft}
\end{figure}

\begin{figure}[!ht]
\centering
\includegraphics[width=\textwidth]{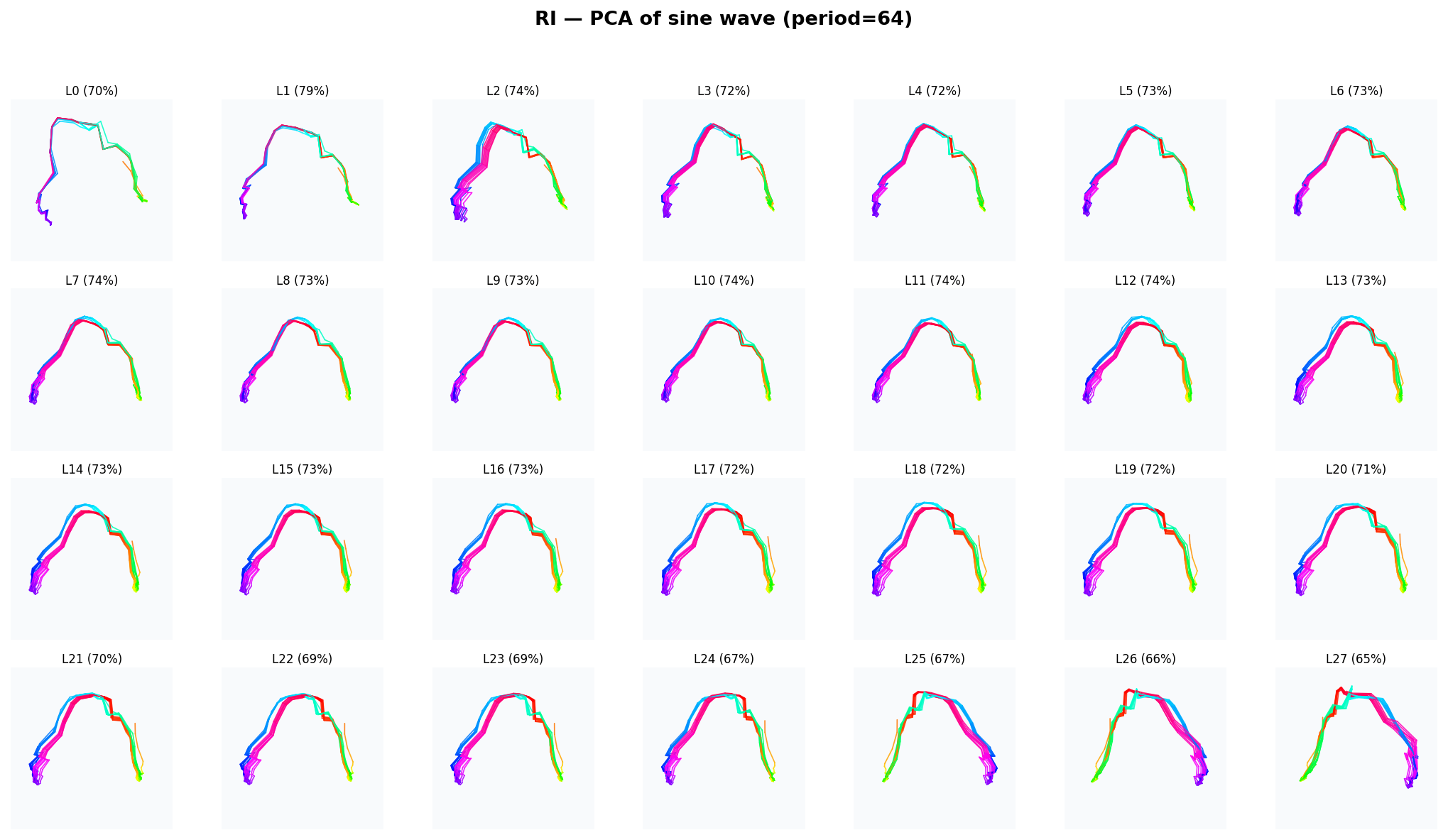}
\caption{\textbf{RandInit: sine wave PCA trajectories across all 28 layers.} Same setup as Figure~\ref{fig:pca_grid_ft}. RandInit produces nearly identical clean arcs at every layer with uniformly high PCA variance (67--79\%), confirming that its learned representation is geometrically homogeneous across the network.}
\label{fig:pca_grid_ri}
\end{figure}

\begin{figure}[!ht]
\centering
\includegraphics[width=\textwidth]{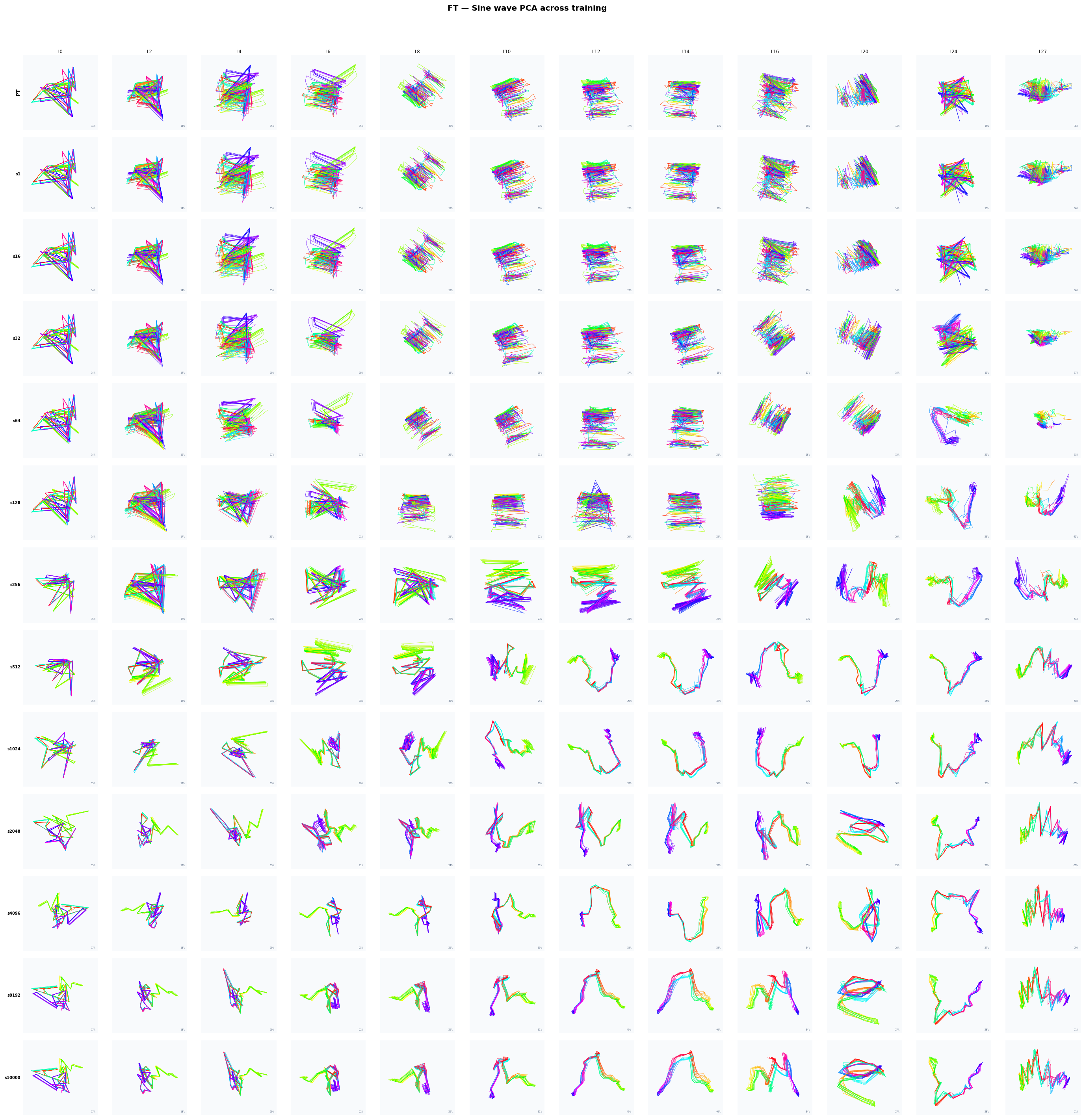}
\caption{\textbf{LangInit: sine wave representations across training checkpoints.} Rows are training steps (top: PT baseline; bottom: final checkpoint at step~10{,}000), columns are selected layers. LangInit begins from the pretrained model's complex trajectories and gradually reshapes them into structured loops, with the transition occurring around steps 256--1024.}
\label{fig:evolution_grid_ft}
\end{figure}

\begin{figure}[!ht]
\centering
\includegraphics[width=\textwidth]{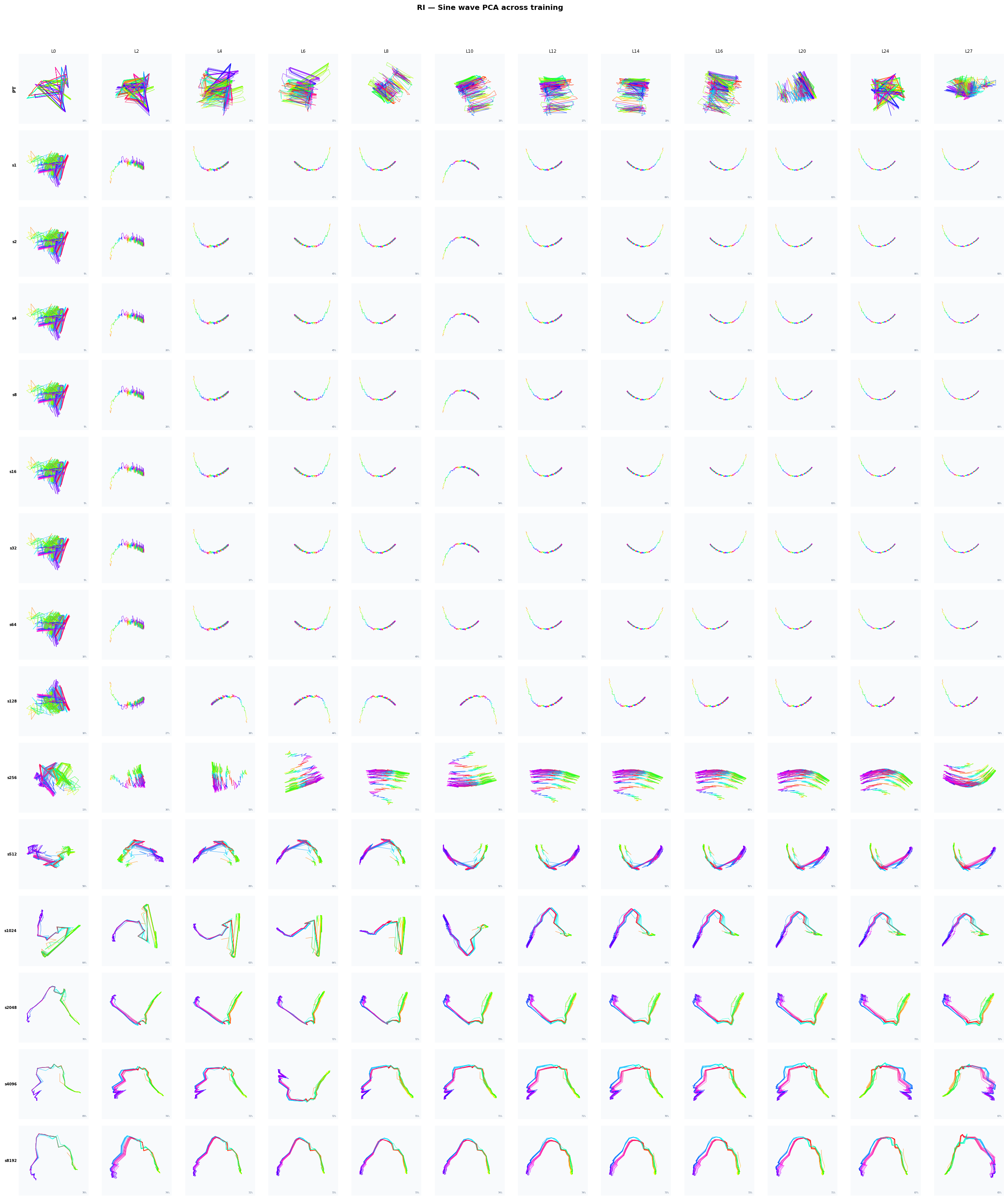}
\caption{\textbf{RandInit: sine wave representations across training checkpoints.} Same setup as Figure~\ref{fig:evolution_grid_ft}. RandInit starts from near-empty representations (random weights) and converges to uniform clean arcs by step~2048, with all layers collapsing to the same geometry simultaneously.}
\label{fig:evolution_grid_ri}
\end{figure}
\clearpage

\section{Cross-Domain Feature Analysis via Crosscoders}
\label{app:crosscoder}

The circuit-level analysis in Appendix~\ref{app:circuit} shows that specific components are shared between time-series periodicity prediction and repetitive language modeling. Here we complement that \emph{component-level} analysis with a \emph{feature-level} analysis using crosscoders---sparse autoencoders with a shared encoder applied across domains---to discover individual latent features that fire on both time-series inputs and semantically coherent natural-language passages.

\subsection{Method}
\label{app:crosscoder_method}

\paragraph{Setup.}
We train one \emph{linear crosscoder} per transformer layer of Qwen3-0.6B.  Each crosscoder has a shared encoder $\mathbf{W}_{\mathrm{enc}} \in \mathbb{R}^{1024 \times 4096}$ followed by Top-$K$ sparsity ($K{=}64$), and two per-domain decoders $\mathbf{W}_{\mathrm{dec}}^{(\mathrm{PT})}, \mathbf{W}_{\mathrm{dec}}^{(\mathrm{FT})} \in \mathbb{R}^{4096 \times 1024}$ (12.6\,M parameters total, float32).  The encoder is applied independently to each domain's hidden states; because the weights are shared, the same latent feature can fire on both pretrained (PT) and finetuned (FT) inputs.

\paragraph{Domains.}
\emph{PT}: Time series formatted as space-separated decimal strings (e.g., \texttt{0.123 -0.456 1.789\,\ldots}), tokenized by the Qwen3-0.6B tokenizer.  Sub-tokens corresponding to each timestep value are mean-pooled to produce one $1024$-dim vector per timestep.
\emph{FT}: The same time series normalized per-window ($z$-score), clipped to $[-5,5]$, and uniformly binned into 1024 tokens for the finetuned model.

\paragraph{Training.}
Input: 3{,}000 windows ($T{=}512$) from GiftEval~\citep{aksu2024giftevalbenchmarkgeneraltime}, precomputed as memory-mapped hidden-state arrays.  Loss: per-domain MSE between normalized inputs and decoder reconstructions, summed over PT and FT.  Dead-feature recovery (AuxK; top-64 among features firing ${<}1$\% over the last 1{,}000 steps, weight $\frac{1}{32}$, active after step 1{,}000).  AdamW, $\mathrm{lr}{=}3{\times}10^{-4}$, 500 warmup steps, cosine decay over 10{,}000 steps.  Early stopping after 1{,}500 steps without validation improvement (minimum step 2{,}000).

\paragraph{Feature analysis pipeline.}
After training, 50{,}000 time-series windows are encoded; for each of the 4{,}096 features we record the PT and FT firing rates (fraction of windows with non-zero activation).  Features firing ${\geq}1$\% in both domains are labeled \textsc{PT\_FT} and ranked by $\text{balance} = \min(\text{rate}_{\mathrm{PT}}, \text{rate}_{\mathrm{FT}})$.  For the top-30 balanced features, we extract the 10 highest-activating time-series windows and 10 highest-activating WikiText-103 passages (30{,}000 sequences, 512 tokens each, encoded through the PT model and the shared crosscoder encoder with separate normalization statistics).

\subsection{Results}
\label{app:crosscoder_results}

Below we present four features with the clearest cross-domain links: each fires on a specific time-series pattern \emph{and} on thematically coherent WikiText passages.

\begin{figure}[h]
\centering
\includegraphics[width=\textwidth]{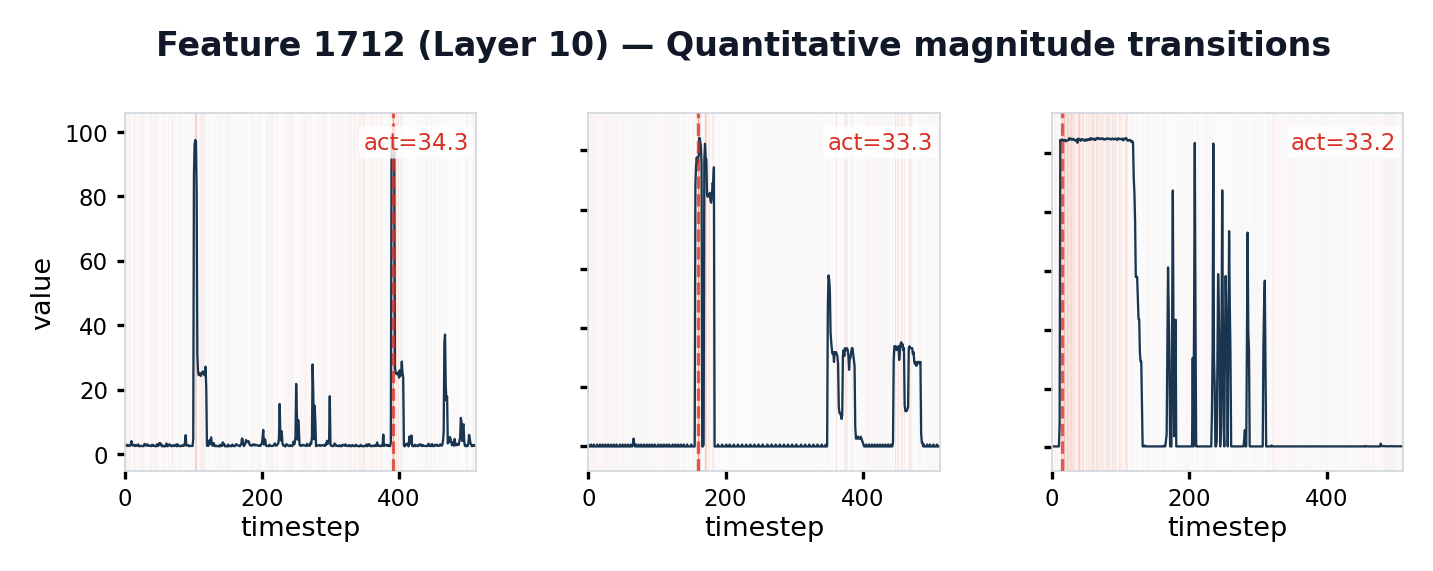}
\caption{\textbf{Feature~1712 (Layer~10): Quantitative magnitude transitions.}  Top-3 activating time-series windows.  Blue: raw signal; red dashed: peak-activation timestep; orange shading: activation intensity.  All three windows share a sudden jump from a low baseline to a high-magnitude plateau.}
\label{fig:crosscoder_magnitude}
\end{figure}

\paragraph{Feature 1712 (Layer~10) --- Quantitative magnitude transitions.}
PT firing rate: 50.4\%, FT: 41.8\%, WikiText: 14.1\%.  This feature detects step changes and regime shifts in time series---values that jump from a near-zero baseline to a sustained high plateau---and fires on text passages dense with numerical measurements and unit conversions (Figure~\ref{fig:crosscoder_magnitude}).

Top-5 WikiText passages at the peak-activation token:
\begin{enumerate}[nosep, leftmargin=1.5em]
\item \emph{(act=14.1)} ``\ldots becoming later in the year by about two days every 243-year cycle.  Transits usually occur in pairs, on nearly the same date eight years apart.''
\item \emph{(act=13.8)} ``In practice, forward premiums and discounts are quoted as annualized percentage deviations from the spot exchange rate\ldots''
\item \emph{(act=13.7)} ``Wages are reflective of the type of jobs available locally, including higher than average employment in manufacturing and the public sector.  The working age population of the town in 2011\ldots''
\item \emph{(act=13.7)} ``So for americium-241, the resistivity at 4.2\,K increases with time from about 2\,\textmu Ohm$\cdot$cm to 10\,\textmu Ohm$\cdot$cm after 40 hours, and saturates at about 16\,\textmu Ohm$\cdot$cm\ldots''
\item \emph{(act=13.7)} ``Falcon's Fury can theoretically accommodate 800 riders per hour.  Carbon-fiber wings buttress each end of a group of seats\ldots''
\end{enumerate}
The shared representation encodes \emph{quantitative magnitude and transition}: literal level shifts in time series, and passages dense with measurements, unit conversions, and numerical comparisons in text.

\begin{figure}[h]
\centering
\includegraphics[width=\textwidth]{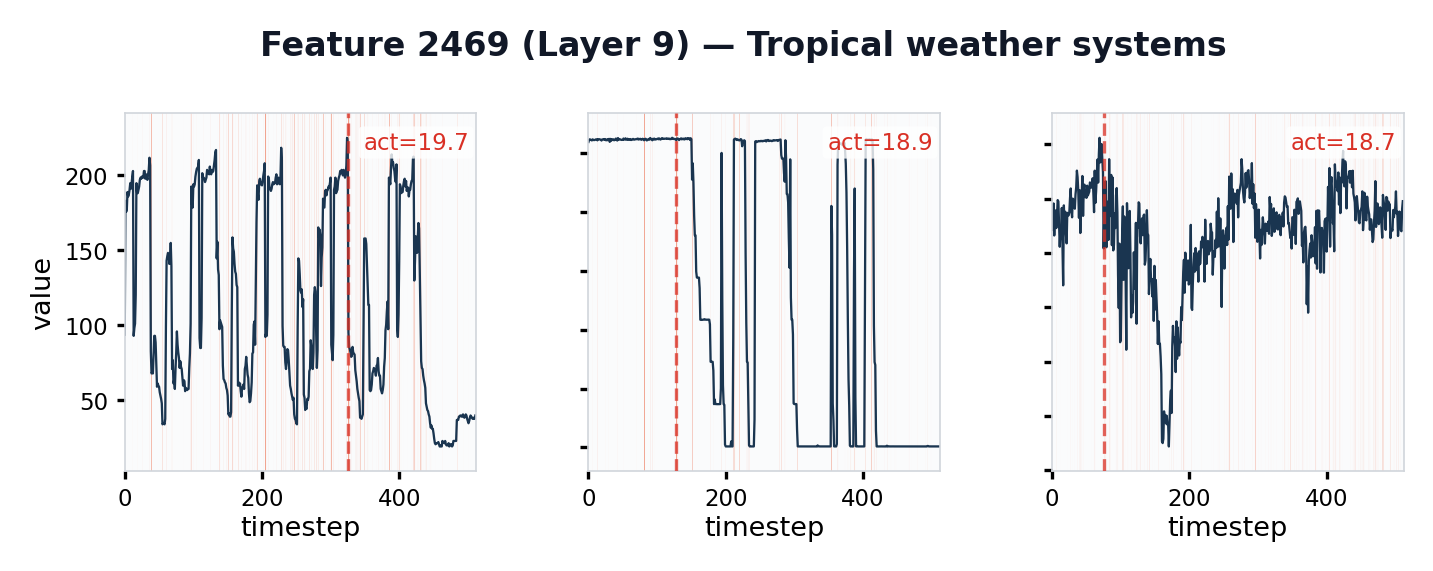}
\caption{\textbf{Feature~2469 (Layer~9): Tropical weather systems.}  Top-3 activating windows.  The left window shows high-variance oscillations with abrupt drops; the middle and right show diverse volatile patterns with regime switching.}
\label{fig:crosscoder_weather}
\end{figure}

\paragraph{Feature 2469 (Layer~9) --- Tropical weather systems.}
PT: 29.9\%, FT: 20.7\%, WikiText: 12.6\%.  Fires on volatile, regime-switching time series and exclusively on tropical cyclone and hurricane narratives in text (Figure~\ref{fig:crosscoder_weather}).

Top-5 WikiText passages:
\begin{enumerate}[nosep, leftmargin=1.5em]
\item \emph{(act=8.7)} ``\ldots the mean locus of formation shifts westward to the Caribbean and Gulf of Mexico, reversing the eastward progression of June through August.  Wind shear from westerlies increases substantially through November\ldots''
\item \emph{(act=8.3)} ``\ldots due to a combination of very high wind shear and dry air.  By October~17, most of the deep convection associated with the system dissipated; however, a brief decrease in wind shear allowed Omar to re-strengthen\ldots''
\item \emph{(act=8.1)} ``The wave continued westward and related thunderstorm activity increased during the following week.  The convective system organized into Tropical Depression Twenty-E on September~28\ldots''
\item \emph{(act=8.1)} ``A tropical wave moved across the northeast Pacific Ocean and formed a tropical depression south of Mexico on October~16.  It strengthened at a moderate pace and reached hurricane intensity on October~18.''
\item \emph{(act=7.9)} ``\ldots formation of Typhoon Chanchu in the western Pacific enhanced convective activity over the Bay of Bengal.  By April~22, a trough developed along an axis from the southern Bay of Bengal eastward to the Andaman Sea.''
\end{enumerate}
The time-series patterns---volatile signals with sudden regime changes---mirror the physical phenomena described in the text: tropical storms that intensify, weaken under wind shear, and shift track.

\begin{figure}[h]
\centering
\includegraphics[width=\textwidth]{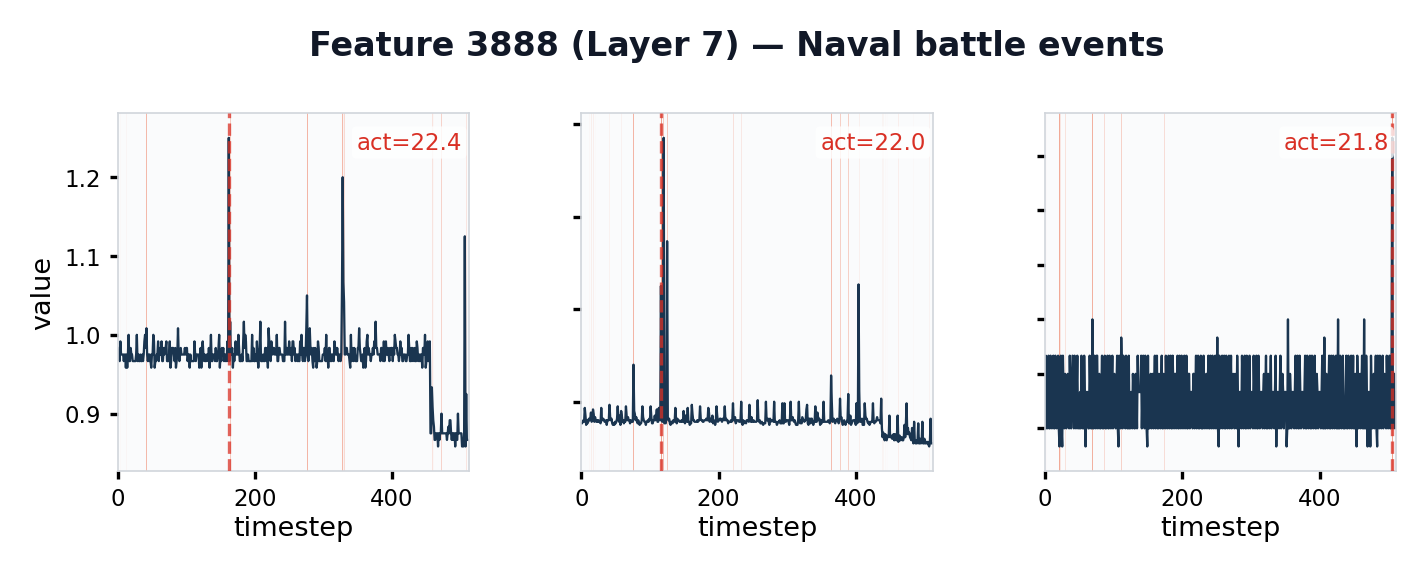}
\caption{\textbf{Feature~3888 (Layer~7): Naval battle events.}  Top-3 activating windows.  Each shows a low-variance baseline punctuated by sharp spikes at the peak-activation timestep (red dashed line).}
\label{fig:crosscoder_naval}
\end{figure}

\paragraph{Feature 3888 (Layer~7) --- Naval battle events.}
PT: 18.5\%, FT: 26.3\%, WikiText: 10.5\%.  Fires on sharp isolated spikes in otherwise stable time series and on naval/military battle narratives with precise timestamps (Figure~\ref{fig:crosscoder_naval}).

Top-5 WikiText passages:
\begin{enumerate}[nosep, leftmargin=1.5em]
\item \emph{(act=10.1)} ``King George V had only 32 percent of her fuel left while Rodney had only enough fuel to continue the chase at high speed until 8:00 the following day.  Admiral Tovey signalled his battlegroup\ldots''
\item \emph{(act=9.3)} ``At 7:20 on 19~July, the destroyer force spotted and was spotted by a pair of Italian light cruisers; Giovanni dalle Bande Nere and Bartolomeo Colleoni, which opened fire seven minutes later.''
\item \emph{(act=9.2)} ``Shortly before 16:00 the battlecruisers of I~Scouting Group encountered the British 1st Battlecruiser Squadron under the command of Vice Admiral David Beatty.  The opposing ships began an artillery\ldots''
\item \emph{(act=9.2)} ``The eastern wind was not communicated to the aircraft, but was 270\textdegree, varying from 20 to 40~knots (37 to 74~km/h).  The take-off started at 14:42:43\ldots''
\item \emph{(act=9.1)} ``\ldots torpedo boat attacks and at 07:30, Burrough sent Eskimo and Somali back to help Manchester but they arrived too late, took on survivors\ldots''
\end{enumerate}
Both modalities encode \emph{sudden, precisely-located events}: an anomalous spike at a single timestep in time series, and a precisely-timestamped combat event in text.

\paragraph{Feature 2567 (Layer~8) --- Missing / null data.}
PT: 35.9\%, FT: 36.8\%, WikiText: 9.4\%.  This feature fires exclusively on NaN/missing time-series data (all 10 top-activating windows are entirely NaN, with peak activation 40.3) and on \texttt{<unk>} tokens and incomplete references in text.

Top-5 WikiText passages:
\begin{enumerate}[nosep, leftmargin=1.5em]
\item \emph{(act=19.8)} ``\ldots at the Royal Navy School of Flight Deck Operations at RNAS Culdrose.  The following is an incomplete list of some of the surviving aircraft.''
\item \emph{(act=18.5)} ``\texttt{<unk>}, \texttt{<unk>}, \texttt{<unk>}, \texttt{<unk>}, \texttt{<unk>}, Ulaid.  Slightly later major groups included the Connachta, \texttt{<unk>}, \texttt{<unk>}.  Smaller groups included the \texttt{<unk>}\ldots''
\item \emph{(act=18.0)} ``\ldots he encountered bad weather, forcing him to return to Japan with heavy damage.  Without waiting for Vizcaino, another ship---built in Izu by the Tokugawa shogunate\ldots''
\item \emph{(act=17.8)} ``Luke 9: \texttt{<unk>}-\texttt{<unk>} --- $\kappa\alpha\grave{\iota}$ \texttt{<unk>}, \texttt{<unk>} \texttt{<unk>} \texttt{<unk>} $\pi\nu\varepsilon\acute{\upsilon}\mu\alpha\tau o\varsigma$ \texttt{<unk>} \texttt{<unk>}\ldots''
\item \emph{(act=17.7)} ``\ldots whom he married in the late 250s when she was 17 or 18 years old.  The number of children Odaenathus had with his first wife is unknown and only one is attested.''
\end{enumerate}
The model represents \emph{absent information} identically across modalities: NaN values in time series and \texttt{<unk>} tokens in text both occupy the same region of representation space.

\section{Causal Circuit Identification}
\label{app:circuit}

The correlational analyses in Section~\ref{sec:finetuning} show that finetuning reuses pretrained directions and that cross-domain features exist. Here we present a preliminary causal analysis: we identify specific model components responsible for periodic time-series prediction and test what role those components play in language modeling.

\subsection{Method}
\label{app:circuit_method}

Since IO-only finetuning only trains embedding and LM-head layers, its 28 transformer layers are identical to PT. We zero-ablate~\citep{wang2024easyediteasytouseknowledgeediting} each of the 476 components (448 attention heads + 28 MLPs) on IO-only finetuning and measure the loss increase ($\Delta\mathcal{L}$) on periodic time series versus non-periodic controls.

\paragraph{Zero-ablation implementation.}
For each attention head, we register a \texttt{forward\_pre\_hook} on the output projection (\texttt{o\_proj}) that zeros the head's 128-dimensional slice of the concatenated 2048-dimensional head output before projection. For each MLP layer, we register a \texttt{forward\_hook} that replaces the MLP output with zeros, so the residual stream passes through unchanged. All ablations are performed under \texttt{torch.no\_grad()}.

\paragraph{Synthetic time series.}
We generate 20 windows (length 512) for each of nine synthetic types. \emph{Periodic}: sine, square wave, sawtooth, seasonal (trend + oscillation), damped sine. \emph{Control}: white noise, constant, linear trend, random walk. Each window is independently $z$-score normalized, clipped to $[-5, 5]$, and uniformly binned into 1024 tokens (matching the IO-only finetuning tokenizer). This yields 180 evaluation windows total (100 periodic, 80 control).

\paragraph{Selectivity metric.}
For each ablated component we compute:
\begin{align}
\Delta\mathcal{L}_\text{periodic} &= \bar{\mathcal{L}}_\text{ablated}^\text{periodic} - \bar{\mathcal{L}}_\text{baseline}^\text{periodic}, \\
\Delta\mathcal{L}_\text{control} &= \bar{\mathcal{L}}_\text{ablated}^\text{control} - \bar{\mathcal{L}}_\text{baseline}^\text{control}, \\
\text{selectivity} &= \Delta\mathcal{L}_\text{periodic} - \Delta\mathcal{L}_\text{control}.
\end{align}
Components with high $\Delta\mathcal{L}_\text{periodic}$ and high selectivity are specifically critical for periodic prediction rather than general-purpose.

\paragraph{Cumulative ablation.}
We compose multiple ablation hooks simultaneously using nested context managers. For the pairwise sweep, we test all $\binom{8}{2} = 28$ pairs of the top-8 selective components. For the growing cumulative, we add components one at a time in order of individual $\Delta\mathcal{L}_\text{periodic}$. The superadditivity ratio is $\rho = \Delta\mathcal{L}_\text{combined} / \sum_i \Delta\mathcal{L}_i^\text{individual}$~\citep{conmy2023automatedcircuitdiscoverymechanistic, elhage2021mathematical}; we use a threshold of $\rho > 1.2$ to account for noise.

\paragraph{WikiText transfer.}
We ablate the top-5 circuit components simultaneously on 2{,}000 WikiText-103 passages (length 512) through the pretrained model (which shares identical transformer layers with IO-only finetuning). For each passage we record the per-sequence cross-entropy loss before and after ablation. The 50 most-degraded and 50 least-degraded passages are extracted for qualitative analysis.

\subsection{Results}

\paragraph{Individual component sweep.}
Table~\ref{tab:ablation_top} shows the top-8 components ranked by $\Delta\mathcal{L}_\text{periodic}$. The two most impactful are both in Layer~1: MLP\textsubscript{L1} ($\Delta\mathcal{L}_p = 5.75$, selectivity $= 3.44$) and head~L1H4 ($\Delta\mathcal{L}_p = 4.50$, selectivity $= 3.81$). Head~L20H0 is the third most impactful with the highest selectivity ($3.28$).

\begin{table}[h]
\centering
\caption{Top-8 components by $\Delta\mathcal{L}$ on periodic TS under zero-ablation. Selectivity $= \Delta\mathcal{L}_\text{periodic} - \Delta\mathcal{L}_\text{control}$ isolates periodicity-specific impact. Full sweep: 476 components.}
\label{tab:ablation_top}
\begin{tabular}{lrrr}
\toprule
Component & $\Delta\mathcal{L}_\text{periodic}$ & $\Delta\mathcal{L}_\text{control}$ & Selectivity \\
\midrule
MLP\textsubscript{L1}   & 5.75 & 2.31 & 3.44 \\
Head L1H4               & 4.50 & 0.69 & 3.81 \\
Head L20H0              & 3.40 & 0.13 & 3.28 \\
Head L13H6              & 3.35 & 2.56 & 0.79 \\
Head L23H6              & 2.00 & 0.06 & 1.94 \\
Head L21H8              & 1.85 & $-0.25$ & 2.10 \\
Head L15H3              & 1.55 & 0.38 & 1.18 \\
Head L11H8              & 1.25 & 0.56 & 0.69 \\
\bottomrule
\end{tabular}
\end{table}

\paragraph{Cumulative ablation reveals a composed circuit.}
The Layer~1 pair (head~L1H4 + MLP\textsubscript{L1}) is strongly superadditive: ablating them together produces $\Delta\mathcal{L}_p = 15.50$, which is 51\% larger than the sum of their individual effects ($\rho = 1.51$; Table~\ref{tab:cumulative}). No other pair among the 28 tested exceeds the $\rho > 1.2$ threshold. Adding head~L20H0 maintains superadditivity ($\rho = 1.36$); beyond three components, returns become subadditive ($\rho < 1$), indicating compensation rather than composition.

\begin{table}[h]
\centering
\caption{Cumulative ablation. The Layer~1 pair is strongly superadditive ($\rho = 1.51$); the core circuit saturates at 3 components.}
\label{tab:cumulative}
\begin{tabular}{lrrrr}
\toprule
Components & $\Delta\mathcal{L}_\text{periodic}$ & $\sum \Delta\mathcal{L}_\text{indiv.}$ & $\rho$ & $\Delta\mathcal{L}_\text{control}$ \\
\midrule
Head L1H4 + MLP\textsubscript{L1} & 15.50 & 10.25 & \textbf{1.51} & 2.38 \\
+ Head L20H0 (top-3)              & 18.55 & 13.65 & \textbf{1.36} & 3.69 \\
+ Head L21H8 (top-4)              & 18.40 & 15.50 & 1.19 & 3.06 \\
+ Head L23H6 (top-5)              & 18.55 & 17.50 & 1.06 & 3.00 \\
All top-8                          & 18.30 & 23.65 & 0.77 & 3.13 \\
\bottomrule
\end{tabular}
\end{table}

\paragraph{The periodicity circuit in language.}
We ablate the top-5 circuit components simultaneously on 2{,}000 WikiText-103 passages through the pretrained model. The circuit ablation causes a mean loss increase of $\Delta\mathcal{L} = 4.53$ nats---two orders of magnitude larger than any individual head's WikiText effect ($\leq 0.04$), confirming circuit-level interaction.

The most-degraded passages ($\Delta\mathcal{L} = 8$--$9$; Table~\ref{tab:wiki_full}) are dominated by sequential enumerations and repeating grammatical templates: biographical sequences (``married to\ldots she gave birth to\ldots''), game mechanics with parallel clause structure (``sets a task for each stage\ldots this task must be completed\ldots the player with the most''), Billboard chart progressions, award category listings, and structured Wikipedia sections with repeating headers. In contrast, the least-degraded passages ($\Delta\mathcal{L} < 2$) are dense, non-repetitive prose: ecclesiastical titles, academic citations, canonical law, and literary criticism---text where predicting the next token depends on semantic content rather than structural repetition.

This suggests the circuit tracks \emph{sequential repetitive structure}---the same abstract property shared by periodic time series and repetitive text. However, the evidence is preliminary: the WikiText passages most degraded by the circuit do not all show an obvious repetitive pattern, and disentangling the circuit's role in general sequence modeling from periodicity-specific computation requires further investigation.

\begin{table}[h]
\centering
\caption{WikiText-103 passages most and least degraded by ablation of the periodicity circuit (top-5 components). Mean $\Delta\mathcal{L} = 4.53$ across 2{,}000 passages.}
\label{tab:wiki_full}
\small
\begin{tabularx}{\textwidth}{r >{\raggedright\arraybackslash}X}
\toprule
$\Delta\mathcal{L}$ & Passage excerpt \\
\midrule
\multicolumn{2}{l}{\emph{Top 20 most degraded}} \\
9.19 & ``\ldots Townsend has been married to Tracy Turner, his girlfriend since he was 19. She gave birth to thei[r]\ldots'' \\
8.88 & ``\ldots Preston winger Will Hayhurst, a Republic of Ireland under-21 international, was signed on a one-month loan\ldots'' \\
8.88 & ``\ldots with the NHK Symphony Orchestra, but cancelled both deals upon Mwanga's return from Japan. Mwanga immediately quit\ldots'' \\
8.88 & ``\ldots his/her final score on the song, with money being awarded in Guitar Hero World Tour. The games have also added\ldots'' \\
8.75 & ``\ldots Viscount, sets a task for each stage. This task must be completed before the player can continue to another map\ldots'' \\
8.50 & ``\ldots The player character is Michel Ardan, an eccentric and intrepid French scientist who is enthusiastic, daring\ldots'' \\
8.38 & ``\ldots Best Music Video, Long Form. In 1998, the categories were retitled Best Short Form Music Video, and Best Long\ldots'' \\
8.31 & ``\ldots on the Billboard Hot 100 twenty-nine, the highest U.S. entry among all singles released from the album\ldots'' \\
8.31 & ``\ldots long run, with The A.V. Club attributing much of the show's early success to the character\ldots'' \\
8.25 & ``\ldots Yankovic recorded `Here's Johnny', a parody of `Who's Johnny' by El DeBarge. The song, a loving ode to\ldots'' \\
8.19 & ``\ldots The music video of `Crazy in Love', released in May 2003, was directed by Jake Nava and filmed\ldots'' \\
8.19 & ``\ldots Finishing with the worst record in the NHL, Columbus had the best chance of receiving the first overall pick\ldots'' \\
8.13 & ``\ldots while one in the Pyramid Texts says the name is based on words shouted by Osiris\ldots'' \\
8.06 & ``\ldots the album Daydream most resembles in its emphasis on R\&B grooves. Tucker specifically complimented `One Sweet Day'\ldots'' \\
8.00 & ``\ldots similar to Konami's Guitar Freaks and to a lesser extent Harmonix's previous music games such as Frequency\ldots'' \\
8.00 & ``\ldots he `hit the wall with play-along music games', and challenged the game makers to explore other ways\ldots'' \\
7.94 & ``\ldots saying that he `was upset. But when you see the talent that was there, it was an honour just to be in the final'\ldots'' \\
7.94 & ``\ldots Flags indicate national team as defined under FIFA eligibility rules. Players may hold more than one non-FIFA\ldots'' \\
7.94 & ``\ldots at the Television Critics Association summer media tour in Beverly Hills, California\ldots'' \\
7.94 & ``\ldots co-wrote and produced a song with Kenneth `Babyface' Edmonds, with whom she had collaborated on Music Box\ldots'' \\
\midrule
\multicolumn{2}{l}{\emph{Top 10 least degraded}} \\
0.83 & ``\ldots Porto e Santa Rufina; Sub-dean of the Sacred College of Cardinals; prefect of the S.C.\ of the Good Government\ldots'' \\
1.48 & ``\ldots Cardinal-Priest of SS.\ Giovanni e Paolo; Grand penitentiary; prefect of the Congregation for the correction\ldots'' \\
1.59 & ``\ldots From the Jewish Question to the Jewish State: An Essay on the Theory of Zionism (thesis), Princeton Univ.\ldots'' \\
1.77 & ``\ldots called Saprang's transfer a `demotion' and a `punishment.' However, Saprang himself claimed that he did not\ldots'' \\
1.82 & ``\ldots the Society of Jesus, provided he observed the canon law; and that it was desirable that the pope should\ldots'' \\
1.89 & ``\ldots the abnormal excess of white blood cells in people with the clinical syndrome described by Velpeau and Bennett\ldots'' \\
1.89 & ``\ldots The protagonist sounds like a `colonial administrator', and his reference to seeking a newer world echoes\ldots'' \\
1.94 & ``\ldots Mbaruk's nephew, Mbaruk bin Rashid, refused to acknowledge the appointment of a new leader\ldots'' \\
1.98 & ``\ldots 00 works were published underground over the course of the war. Literary discussions were held\ldots'' \\
2.02 & ``\ldots Although the poem was defended by a few critics, E.C. Pettet returned to the argument that the poem lacked\ldots'' \\
\bottomrule
\end{tabularx}
\end{table}

\end{document}